\documentclass[sensors,article,accept,pdftex,moreauthors]{Definitions/mdpi} 

\firstpage{1} 
\pubvolume{23}
\issuenum{1}
\articlenumber{9679}
\pubyear{2023}
\copyrightyear{2023}
\externaleditor{Academic Editors:~Jun~Wu, Zhaojun~Steven~Li, Yi~Qin and Carman~K.M.~Lee}
\datereceived{24 August 2022} 
\daterevised{27 November 2023} 
\dateaccepted{2 December 2023} 
\datepublished{7 December 2023} 
\hreflink{https://doi.org/10.3390/s23249679} 
\doinum{10.3390/s23249679}
\pdfoutput=1 

\usepackage{enumerate}
\usepackage{subfiles}
\usepackage[nottoc]{tocbibind}
\usepackage{lipsum}
\usepackage{mathtools}
\usepackage[noend]{algpseudocode}
\usepackage{enumitem,kantlipsum}
\usepackage{indentfirst}
\usepackage{multirow}
\usepackage{pifont}
\usepackage{amsmath,amssymb,amsfonts}
\usepackage{algorithm}
\usepackage{graphicx}
\usepackage{textcomp}
\usepackage{bbding}
\usepackage{rotating}
\usepackage[labelformat=simple]{subcaption} 

\DeclareCaptionLabelFormat{subcaptionlabel}{\normalfont(\textbf{#2}\normalfont)}
\usepackage[bottom,flushmargin]{footmisc}
\usepackage{adjustbox}
\newcommand\footnoteref[1]{\protected@xdef\@thefnmark{\ref{#1}}\@footnotemark}

\Title{Spatio-Temporal Anomaly Detection with Graph Networks for Data Quality Monitoring of the Hadron Calorimeter}

\TitleCitation{Spatio-Temporal Anomaly Detection with Graph Networks for Data Quality Monitoring of the Hadron Calorimeter}

\Author{Mulugeta~Weldezgina~Asres~$^{1,}$*\orcidA{}, Christian~Walter~Omlin~$^{1,}$*\orcidB{}, Long~Wang~$^{2}$\orcidC{}, David~Yu~$^{3}$\orcidD{}, Pavel~Parygin~$^{4}$\orcidE{}, \mbox{Jay~Dittmann~$^{5}$\orcidF{}}, Georgia~Karapostoli~$^{6}$\orcidG{}, Markus~Seidel~$^{7}$\orcidH{}, Rosamaria~Venditti~$^{8}$\orcidI{}, Luka~Lambrecht~$^{9}$\orcidJ{}, Emanuele~Usai~$^{10}$\orcidK{}, Muhammad~Ahmad~$^{11}$\orcidL{}, Javier~Fernandez~Menendez~$^{12}$\orcidM{}, Kaori~Maeshima~$^{13}$\orcidN{} and~the~CMS-HCAL {Collaboration} 
 $^\dagger$}

\AuthorNames{Mulugeta Weldezgina Asres, Christian Walter Omlin, Long Wang, David Yu, Pavel Parygin, Jay Dittmann, 
Georgia Karapostoli, Markus Seidel, Rosamaria Venditti, Luka Lambrecht, Emanuele Usai, Muhammad Ahmad, Javier Fernandez Menendez, Kaori Maeshima and the CMS-HCAL Collaboration.
}

\AuthorCitation{Asres, M.W.; Omlin, C.W.; Wang, L.; Yu, D.; Parygin, P; Dittmann, J.; Karapostoli, G.; Seidel, M.; Venditti, R.; Lambrecht, L.; et al.
}
\address{$^{1}$ \quad {Centre~for~Artificial~Intelligence Research}, {Department~of~Information~and~Communication~Technology, University~of~Agder, 4879~Grimstad,~Norway} \\
$^{2}$ \quad {Department~of~Physics}, {University~of~Maryland}, {College~Park}, {MD~20742},~{USA}; long.wang@cern.ch \\
$^{3}$ \quad {Department~of~Physics}, Brown~University, {Providence},~{RI~02912},~{USA}; dryu@fnal.gov \\
$^{4}$ \quad {Department~of~Physics~and~Astronomy},  {University~of~Rochester}, {Rochester},~{NY~14627},~{USA}; pavel.parygin@cern.ch \\
$^{5}$ \quad {Department~of~Physics}, Baylor~University, {Waco},~{TX~76706},~{USA}; jay\_dittmann@baylor.edu \\
$^{6}$ \quad {Department~of~Physics~\&~Astronomy}, University~of~California, {Riverside},~{CA~92521},~{USA}; georgia.karapostoli@cern.ch \\
$^{7}$ \quad {Institute~of~Particle~Physics~and~Accelerator~Technologies}, Riga~Technical~University, {Rīga}~{LV-1048},~{Latvia}; markus.seidel@cern.ch \\
$^{8}$ \quad {Department~of~Physics}, {Bari~University}, {70121~Bari},~{Italy}; rosamaria.venditti@cern.ch \\
$^{9}$ \quad {Department~of~Physics~and~Astronomy,} Ghent~University, {B-9000~Ghent},~{Belgium}; luka.lambrecht@cern.ch \\
$^{10}$ \quad {Department~of~Physics~and~Astronomy}, University~of~Alabama, {Tuscaloosa},~{AL~35487},~{USA}; eusai@ua.edu \\
$^{11}$ \quad {Department~of~Physics~and~Astronomy}, Texas~A\&M~University, {College~Station},~{TX~77843},~{USA}; m.ahmad@cern.ch \\
$^{12}$ \quad {Instituto~Universitario~de~Ciencias~y~Tecnologías Espaciales~de~Asturias}, {University~of~Oviedo}, {33004~Oviedo},~{Spain}; javier.fernandez.menendez@cern.ch \\
$^{13}$ \quad Fermi~National~Accelerator~Laboratory,~{Batavia},~{IL~60510},~{USA}; kaori.maeshima@cern.ch \\
}

\corres{Correspondence: mulugetawa@uia.no (M.W.A.); christian.omlin@uia.no (C.W.O.)}
\firstnote{The CMS-HCAL Collaboration {author list is} given at the {Supplemental File.} 
}

\abstract{
The Compact Muon Solenoid (CMS) experiment is a general-purpose detector for high-energy collision at the Large Hadron Collider (LHC) at CERN. It employs an online data quality monitoring (DQM) system to promptly spot and diagnose particle data acquisition problems to avoid data quality loss.  
In this study, we present a semi-supervised spatio-temporal anomaly detection (AD) monitoring system for the physics particle reading channels of the Hadron Calorimeter (HCAL) of the CMS using three-dimensional digi-occupancy map data of the DQM. We propose the GraphSTAD system, which employs convolutional and graph neural networks to learn local spatial characteristics induced by particles traversing the detector and the global behavior owing to shared backend circuit connections and housing boxes of the channels, respectively. Recurrent neural networks capture the temporal evolution of the extracted spatial features. We validate the accuracy of the proposed AD system in capturing diverse channel fault types using the LHC collision data sets. The GraphSTAD system achieves production-level accuracy and is being integrated into the CMS core production system for real-time monitoring of the HCAL. We provide a quantitative performance comparison with alternative benchmark models to demonstrate the promising leverage of the presented system.
Code: \href{https://github.com/muleina/CMS_HCAL_ML_OnlineDQM}{https://github.com/muleina/CMS\_HCAL\_ML\_OnlineDQM}
}

\keyword{anomaly detection; monitoring; spatio-temporal; deep learning; graph networks; particle sensors; CMS; LHC} 

\begin{document}

\section{Introduction}
\label{sec:introduction}

Deep learning (DL) has become increasingly prevalent for anomaly detection (AD) applications for reliability, safety, and health monitoring in several domains with the proliferation of sensor data in recent years~\cite{chalapathy2019deep, zhao2022comparative, cook2019anomaly}. AD has been applied for a diverse set of tasks, including but not limited to machinery fault diagnosis and prognosis~\cite{li2020intelligent, zhou2021harnessing}, electronic device fault diagnosis~\cite{shi2019open, wielgosz2018model, wielgosz2017using, asres2021unsupervised}, medical diagnosis~\cite{ahmedt2021graph, bakator2018deep, zhou2019beatgan, cowton2018combined}, cybersecurity~\cite{bankovic2012detecting, tivsljaric2021spatiotemporal, kim2019anomaly}, crowd monitoring~\cite{xu2017detecting, chang2022video, luo2019video, wu2020fast, hasan2016learning, ullah2021efficient, hu2019efficient}, traffic monitoring~\cite{hsu2017anomaly, deng2022graph}, environment monitoring~\cite{zhang2020improving}, the Internet of things~\cite{cook2019anomaly, liu2020deep}, and energy and power management~\cite{buzau2019hybrid, choi2020gan}. AD aims to determine anomalies depending on the setting and application domain~\cite{zhao2022comparative}.
An anomaly is generally an odd observation---abnormalities, deviants, outliers, discords, failures, intrusions, exceptions, aberrations, peculiarities, or contaminants---from a bulk of observations often indicating peculiar underlying incidents~\cite{chalapathy2019deep}. AD methods can be categorized as supervised or unsupervised approaches: (1) supervised approaches require annotated ground-truth anomaly observations, and (2) unsupervised methods do not require labeled anomaly data and are more generally pragmatic in many real-world application settings, as data annotation is expensive. Unsupervised AD models trained with only healthy observations are often categorized as semi-supervised approaches.

Deep learning has become effective for AD modeling because of its capability to capture complex structures, extract end-to-end automatic features, and scale for large data sets~\cite{chalapathy2019deep, zhao2022comparative}.
Several DL models have been proposed in the literature for diverse data types, such as structural~\cite{chalapathy2019deep}, time series
\cite{asres2021unsupervised, zhao2020multivariate, cowton2018combined, guo2018multidimensional, zhang2019deep, munir2018deepant, canizo2019multi, liu2020deep, kim2019anomaly, niu2020lstm, choi2020gan, li2018anomaly, zhou2019beatgan, deng2021hifi, deng2021graph, wielgosz2017using, wielgosz2018model}, image~\cite{zhang2020improving, ahmedt2021graph}, graph network data~\cite{hsu2017anomaly, deng2022graph, bankovic2012detecting, tivsljaric2021spatiotemporal, jiang2022anomaly}, and spatio-temporal~\cite{hsu2017anomaly, deng2022graph, bankovic2012detecting, tivsljaric2021spatiotemporal, xu2017detecting, chang2022video, luo2019video, wu2020fast, hasan2016learning, ullah2021efficient, ahmedt2021graph, jiang2022anomaly}. 
Spatio-temporal (ST) data are commonly collected in diverse domains, such as visual streaming data~\cite{xu2017detecting, chang2022video, luo2019video, wu2020fast, hasan2016learning, ullah2021efficient, hu2019efficient}, transportation traffic flows~\cite{hsu2017anomaly, deng2022graph}, sensor networks~\cite{bankovic2012detecting, tivsljaric2021spatiotemporal, jiang2022anomaly}, geoscience~\cite{zhang2020improving}, medical diagnosis~\cite{ahmedt2021graph}, and high-energy physics~\cite{collaboration2008cms, duarte2022graph}. A unique quality of ST data that differentiates it from other classic data is the presence of dependencies among measurements induced by the spatial and temporal attributes, where data correlations are more complex to capture by conventional techniques~\cite{atluri2018spatio}. 
ST anomaly is thus defined as a data point or cluster of data points that violate the nominal ST correlation structure of the healthy ST data~\cite{hsu2017anomaly, deng2022graph, bankovic2012detecting, tivsljaric2021spatiotemporal, xu2017detecting, chang2022video, luo2019video, wu2020fast, hasan2016learning, ullah2021efficient, ahmedt2021graph, jiang2022anomaly}. 
The wide range of unsupervised DL AD methods discover anomalies in temporal context using density clustering on latent space~\cite{asres2021unsupervised}, data reconstruction~\cite{asres2021unsupervised, cowton2018combined, zhao2020multivariate}, and prediction~\cite{munir2018deepant, zhao2020multivariate, canizo2019multi, liu2020deep, kim2019anomaly}. Variants of recurrent neural networks (RNNs)~\cite{cowton2018combined, liu2020deep, niu2020lstm, wielgosz2018model, wielgosz2017using, canizo2019multi, asres2021unsupervised, hasan2016learning, hsu2017anomaly, ullah2021efficient}, convolutional neural networks (CNNs)~\cite{munir2018deepant, zhao2020multivariate, liu2020deep, canizo2019multi, asres2021unsupervised, chang2022video, luo2019video, wu2020fast, hasan2016learning}, generative adversarial networks (GANs)~\cite{niu2020lstm, choi2020gan, li2018anomaly, zhou2019beatgan}, graph neural networks (GNNs)~\cite{zhao2020multivariate, deng2021graph, deng2021hifi, hsu2017anomaly, deng2022graph}, and transformers~\cite{deng2021hifi} have been explored and achieved competitive performance for multivariate temporal or ST AD.

The \textit{Large Hadron Collider} (LHC) is the largest particle collider ever built globally. 
It is designed to conduct experiments in physics and increase our understanding of the universe, expecting that new findings will lead to practical applications.
The LHC is a two-ring superconducting hadron accelerator and collider capable of accelerating and colliding beams of protons and heavy ions with the unprecedented luminosity of $10^{34}$ cm$^{-2}$s$^{-1}$ and $10^{27}$ cm$^{-2}$s$^{-1}$, respectively, at a velocity close to the speed of light---$3 \times 10^8$ ms$^{-1}$~\cite{evans2008lhc, heuer2012future}. 
The \textit{Compact Muon Solenoid} (CMS) experiment is a general-purpose detector for \textit{high-energy physics} (HEP) at the LHC~\cite{collaboration2008cms}. 
The CMS employs a \textit{data quality monitoring} (DQM) system to guarantee high-quality physics data through online monitoring that provides live feedback during data acquisition and offline monitoring that certifies the data quality after offline processing~\cite{azzolini2019data}. 
The online DQM identifies emerging problems using a reference distribution and predefined tests to detect known failure modes using summary histograms, such as a digi-occupancy map of the CMS calorimeters~\cite{tuura2010cms, de2014cms}. 
A \textit{digi-occupancy map} contains a histogram record of particle hits of the data-recording channels of the calorimeters. 
The calorimeters could have several flaws, such as issues with the front-end particle sensing scintillators, digitization and communication systems, backend hardware, and algorithms, which are usually reflected in the digi-occupancy map.  
The growing complexity of the detector and the physics experiments make data-driven AD systems essential tools for the CMS to identify and localize detector anomalies automatically. 
{The CMS detector consists of a tracker to reconstruct particle paths accurately, two calorimeters---the \textit{electromagnetic} (ECAL) and the \textit{hadronic} (HCAL) calorimeters to detect electrons, photons, and hadrons, respectively---and several \textit{muon} detectors.} 
The synergy in AD has thus far achieved promising results on spatial 2D histogram maps of the DQM for the ECAL~\cite{azzolin2019improving} and the muon detectors~\cite{pol2019detector}. 

{Previous} studies only considered extreme anomalies, such as no reading, dead, and high-noise, hot-particle-sensing calorimeter channels. 
Detecting degrading channels is essential for quality deterioration monitoring and early intervention, but they are often challenging to capture; for instance, the improperly tuned bias voltage on the HCAL physics-particle-sensing channels caused nonuniformity in the hit map of the DQM, but the channels were neither dead nor hot~\cite{oleksandr2021}. 
The calorimeter channels may degrade with a subtle abnormality before reaching extreme channel fault status.
Capturing such subtle anomalies, e.g., a slow system degradation, makes temporal AD models appealing for early anomaly prediction before ultimate system failure. 
Time-aware models extract temporal context to enhance AD performance. 
A few efforts have thus far been focused on temporal models despite the acknowledged potential in the future automation technology challenges at the LHC~\cite{azzolin2019improving, wielgosz2018model}.
Our study focuses on DQM automation through time-aware AD modeling using digi-occupancy histogram maps of the HCAL.
The digi-occupancy data of the HCAL are 3D due to its depthwise calorimeter segmentation. It poses multidimensional challenges, and it is relatively unexplored in ML endeavors. 
The particle hit map data of the HCAL are highly dependent on the collision luminosity---a measure of how many collisions are happening in a particle accelerator---and the number of particles traversing the calorimeter. 
The effort on data normalization that enhances the learning generalization of machine learning models is still limited.

In this study, we address the above gaps while investigating the performance of temporal DL models in enhancing AD for the HCAL DQM system.
We propose to detect anomalies of the HCAL particle-sensing channels through a semi-supervised AD system---GraphSTAD---from spatial digi-occupancy maps of the DQM. 
Anomalies can be unpredictable and come in different patterns of severity, shape, and size, often limiting the availability of labeled anomaly data covering all possible faults. 
We employ a semi-supervised approach for the AD system; the concept for the AD is that an autoencoder (AE) trained to reconstruct healthy digi-occupancy maps would adequately reconstruct the healthy maps, whereas it would yield a high reconstruction error for maps with anomalies.  
Since abnormal events can have a spatial appearance and temporal context, we combine both the spatial and temporal features---spatio-temporal---for the AD~\cite{xu2017detecting, chang2022video, luo2019video, hasan2016learning, wu2020fast, hsu2017anomaly, ullah2021efficient, hu2019efficient, bankovic2012detecting, zhang2020improving}.
The spatial nature of the digi-occupancy map of the HCAL may exhibit irregularity; although adjacent channels with the Euclidean distance are exposed to collision article hits around their region, the channels may belong to different backend circuits, resulting in a non-Euclidean spatial behavior on the digi measurements.
The GraphSTAD system captures the behavior of channels from regional collision particle hits, and electrical and environmental characteristics due to the shared backend circuit of the channels to effectively detect the degradation of faulty channels. The AD system attains these utilities using a deep AE model that learns the local spatial behavior, the physical-connectivity-induced shared behavior, and the temporal behavior through convolutional neural networks (CNNs), graph neural networks (GNNs), and recurrent neural networks (RNNs), respectively.

We evaluate our proposed AD approach in detecting spatial faults and temporal discords on digi-occupancy maps of the HCAL. We simulate different realistic types of anomalies---\textit{dead channels} without registered hits and \textit{hot channels} dominated by electronic noise---resulting in a much higher hit count than expected, and \textit{degraded channels} with deteriorated particle detection efficiency, resulting in lower hit counts than expected, to analyze the effectiveness of the AD model. The results demonstrate promising performance in detecting and localizing the anomalies. We further validate the efficacy in detecting real anomalies and discuss comparisons to benchmark models and the existing DQM system.

We briefly describe the DQM and HCAL systems in Section \ref{sec:background}, and our data sets in Section \ref{sec:datasetdescription}. Section \ref{sec:methodology} explains the methodology of the proposed GraphSTAD model, and Section \ref{sec:resultsanddiscussion} presents the performance evaluation and result discussion. Finally, we summarize the contribution of our study in Section \ref{sec:conclusion}.

\section{Background}
\label{sec:background}

This section describes the DQM and HCAL systems of the CMS experiment.

\subsection{Data Quality Monitoring of the CMS Experiment}
\label{sec:hcal_dqm_rbx}

The detector and collision data's offline processing complexity requires continuous data quality monitoring. Shifters and physicists at the CMS monitor the collision quality and select data usable for analysis; they look for unexpected issues that could affect the data quality, e.g., noise spikes, dead areas of the detector, and calibration problems~\cite{Chahal2019ippp}. 
The DQM provides feedback on detector performance and data reconstruction; it generates a list of certified data for physics analyses---the ``Golden JSON''~\cite{azzolini2019data}. 
The DQM employs online and offline monitoring mechanisms:
(1) the \textit{online monitoring} is a real-time DQM during data acquisition, and (2) the \textit{offline monitoring}---after 48 h since the collisions were recorded---provides the final fine-grained data quality analysis for data certification. 
The online DQM populates a set of histogram-based maps on a selection of events and provides summary plots with alarms that DQM experts inspect to spot problems. 
The \textit{digi-occupancy maps}---one of the maps generated by the online DQM---incorporate particle hit histogram records of the particle readout channel sensor of the calorimeters. 
A digi---also called hit---is a reconstructed and calibrated collision physics signal of the calorimeter.
Various faults in the calorimeter affecting the front-end hardware and software components appear in the digi-occupancy map. 
Previous efforts by~\cite{azzolin2019improving, azzolini2019data, pol2019anomaly, pol2019detector} demonstrate the promising AD efficacy of using digi-occupancy maps for calorimeter channel monitoring using machine learning.
However, end-to-end DL with temporal models is relatively unexplored~\cite{azzolin2019improving, pol2019detector}.

The purpose of leveraging the DQM through machine learning is to address particular challenges: (1) the latency of human intervention and thresholds require sufficient statistics; (2) the volume of data a human can process in a finite time is limited; (3) rule-based approaches do not scale and assume limited potential failure scenarios; (4) dynamic running conditions change reference samples; (5) the effort to train human shifters who monitor DQM dashboards and maintain instructions is expensive. 
Developing machine learning models for the DQM comes with some impediments despite the potential promises; data normalization to handle variation in experimental settings, the granularity of the failures to spot, and limited availability of the ground-truth labels are among the challenges~\cite{pol2019detector}.

We extend the efforts in AD with ST modeling of the digi-occupancy maps of the DQM for the HCAL.
Several promising ST AD models have been proposed in the literature in diverse domains~\cite{xu2017detecting, chang2022video, luo2019video, wu2020fast, hasan2016learning, ullah2021efficient, hu2019efficient, hsu2017anomaly, deng2022graph, bankovic2012detecting, tivsljaric2021spatiotemporal, jiang2022anomaly, zhang2020improving, ahmedt2021graph}. 
The previous AD studies on video data sets~\cite{chang2022video, luo2019video, wu2020fast, hasan2016learning} focus on CNNs for regular spatial feature extraction, and GNNs are gaining popularity for sensor and traffic flow data~\cite{hsu2017anomaly, deng2022graph} that exhibit irregular spatial attributes with a non-Euclidean distance among nodes. 
GNNs have recently achieved promising results at the LHC~\cite{duarte2022graph, shlomi2020graph} and outperformed CNN in learning irregular calorimeter geometry~\cite{qasim2019learning} and in pileup mitigation~\cite{martinez2019pileup}.
The spatial characteristics of the HCAL channels exhibit a regular spatial positioning of particle hits in the calorimeter and an irregularity in measurement due to adjacent channels may share different backend circuits. 
Our proposed study presents an AD model for the DQM by integrating both CNNs and GNNs~\cite{bruna2013spectral, kipf2016semi} to capture Euclidean and non-Euclidean spatial characteristics, respectively, and an RNN for temporal learning.

\subsection{Readout Boxes of the Hadron Calorimeter}

\begin{figure}[H]
\includegraphics[width=0.485\linewidth]{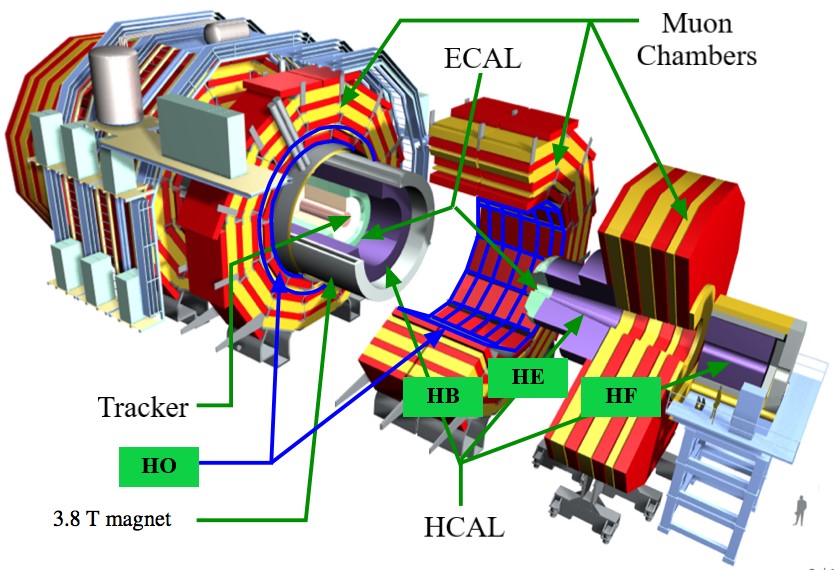}
\caption{Schematic of the CMS detector and its calorimeters~\cite{focardi2012status}.}
\label{fig:cms_diagram_hcal}
\end{figure}

The HCAL is a specialized calorimeter to capture hadronic particles. 
The calorimeter is composed of multiple subsystems such as \textit{HCAL {Endcap}} (HE), \textit{HCAL {Barrel}} (HB), \textit{HCAL {Forward}} (HF), and \textit{HCAL {Outer}} (HO) (see Figure \ref{fig:cms_diagram_hcal}).

\begin{figure}[H]
\includegraphics[width=0.7\linewidth]{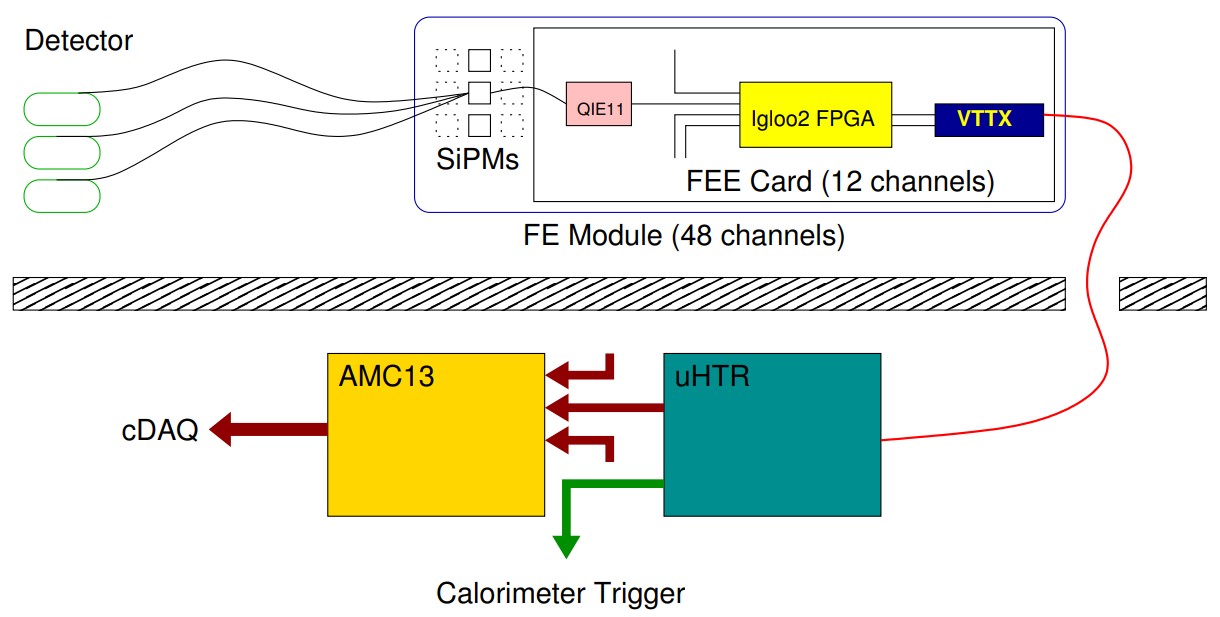}
\caption{The data acquisition chain of the HE, including the SiPMs, the front-end readout cards, and the optical link connected to the backend electronics~\cite{strobbe2017upgrade}. Each readout card contains 10--12 QIE11 for charge integration, an Igloo2 FPGA for data serialization and encoding, and a VTTx optical transmitter. A fault in the chain may cause anomalous digi-occupancy reading in the online DQM.}
\label{fig:HE_data_acquisition_system_chain}
\end{figure}

The HCAL subsystems are made of \textit{readout boxes} (RBXes) to house the data acquisition electronics. 
The RBXes provide high voltage, low voltage, backplane communications, and cooling to the data acquisition electronics. 
The HE---the use-case of our study---consists of 36 RBXes arranged on the plus and minus hemispheres of the CMS. Its front-end particle detection system is built on brass and plastic scintillators, and the produced photon is transmitted via the wavelength-shifting fibers to \textit{silicon photomultipliers} (SiPMs) (see \mbox{Figure \ref{fig:HE_data_acquisition_system_chain}}).
Each RBX comprises 4 \textit{readout modules} (RMs) for signal digitization~\cite{strobbe2017upgrade}; each RM has 48 SiPMs and 4 readout cards, each including 12 \textit{charge-integrating and -encoding} channels (QIE11 ASICs) connected to corresponding SiPMs and a \textit{field-programmable gate array} (Microsemi Igloo2 FPGA). 
A QIE integrates the charge from a SiPM at 40 MHz, and the FPGA serializes and encodes the data from 12 QIE channels (see Figure \ref{fig:HE_data_acquisition_system_chain}). The encoded data are optically transmitted to the backend system via the CERN \textit{versatile twin transmitter} (VTTx) at 4.8 Gbps.
The HE system has 17 detector scintillator layers that are read out in seven groups---hereafter referred to as $depths$; the light from the scintillators in any given group is optically added together by sending it to a single SiPM. Additional channels enable a more refined depth segmentation, ideal for precisely calibrating the depth-dependent radiation damage on the HCAL~\cite{azzolini2019data}.

\section{Data Set Description}
\label{sec:datasetdescription}

We employed digi-occupancy map data of the online DQM system to train and validate the proposed AD system.
The collision data of the LHC are aggregated into runs, each containing thousands of lumisections. A lumisection (LS) corresponds to approximately 23 s of data acquisition and comprises hundreds or thousands of collision events containing particle hit records. 
The digi-occupancy maps generated by the online DQM contain particle hit histogram records of the particle readout channel sensor of the calorimeters. 
Several faults in the calorimeter affecting the front-end particle-sensing scintillators, the digitalization and communication systems, the backend hardware, and the algorithms usually appear in the digi-occupancy map. 
The value of the digi-occupancy varies with the received luminosity recorded by the CMS---hereafter referred to as the luminosity---and the number of events---particles traversing the calorimeter. 
The maps from a sequence of LSs constitute the attribution of ST data with correlated spatial and temporal relations~\cite{atluri2018spatio}.

\begin{figure}[H]
\includegraphics[width=0.4\linewidth]{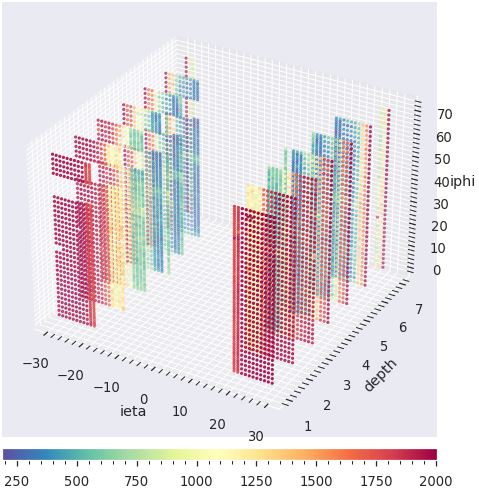}
\caption{Digi-occupancy map (\textit{year = 2018, RunId = 325170, LS = 15}) of the HE. The HE channels are placed in $|i\eta|=[16, 29]$, $i\phi=[1, 72]$, and $depth=[1, 7]$. 
Each pixel in the map corresponds to the readout of an HE channel. The HCAL covers a considerable volume of CMS and has a fine segmentation along three axes ($i\eta$, $i\phi$, and $depth$). 
The missing section at the top left is due to two failed RBXes during the 2018 collision runs.}
\label{fig:he_digioccupancy_sample}
\end{figure}

The digi-occupancy map root-file data sets were collected in 2018 during the LHC RUN-2 collision by the CMS experiment. The data set, from the CMS \textit{ZeroBias Primary Dataset}, contains approximately {20,000} LSs from 20 different healthy runs prescrutinized by the CMS certifiers and declared in the ``Golden JSON'' of the DQM as of good quality~\cite{rapsevicius2011cms}.
The digi-occupancy map data of the HCAL have 3D spatial dimensions with $\eta$ $\phi$, and $depth$ axes and contain digi histogram records of the physics readout channel sensor of the calorimeter referenced by $i\eta=[-32, 32]$, $i\phi=[1, 72]$, and $depth=[1, 7]$ axes (see Figure \ref{fig:he_digioccupancy_sample}). 
The maps---one per LS---were populated with the per-LS received luminosity up to 0.4 $pb^{-1}$ and the number of events up to 2250.
Our working data set contains about {20,000} map samples, each with a dimension of $[i\eta=64 \times i\phi=72 \times depth=7])$.

\section{Methodology}
\label{sec:methodology}

This section presents the proposed GraphSTAD approach for HCAL monitoring using digi-occupancy maps.

\begin{figure}[H]
\includegraphics[width=1\textwidth, scale=1]{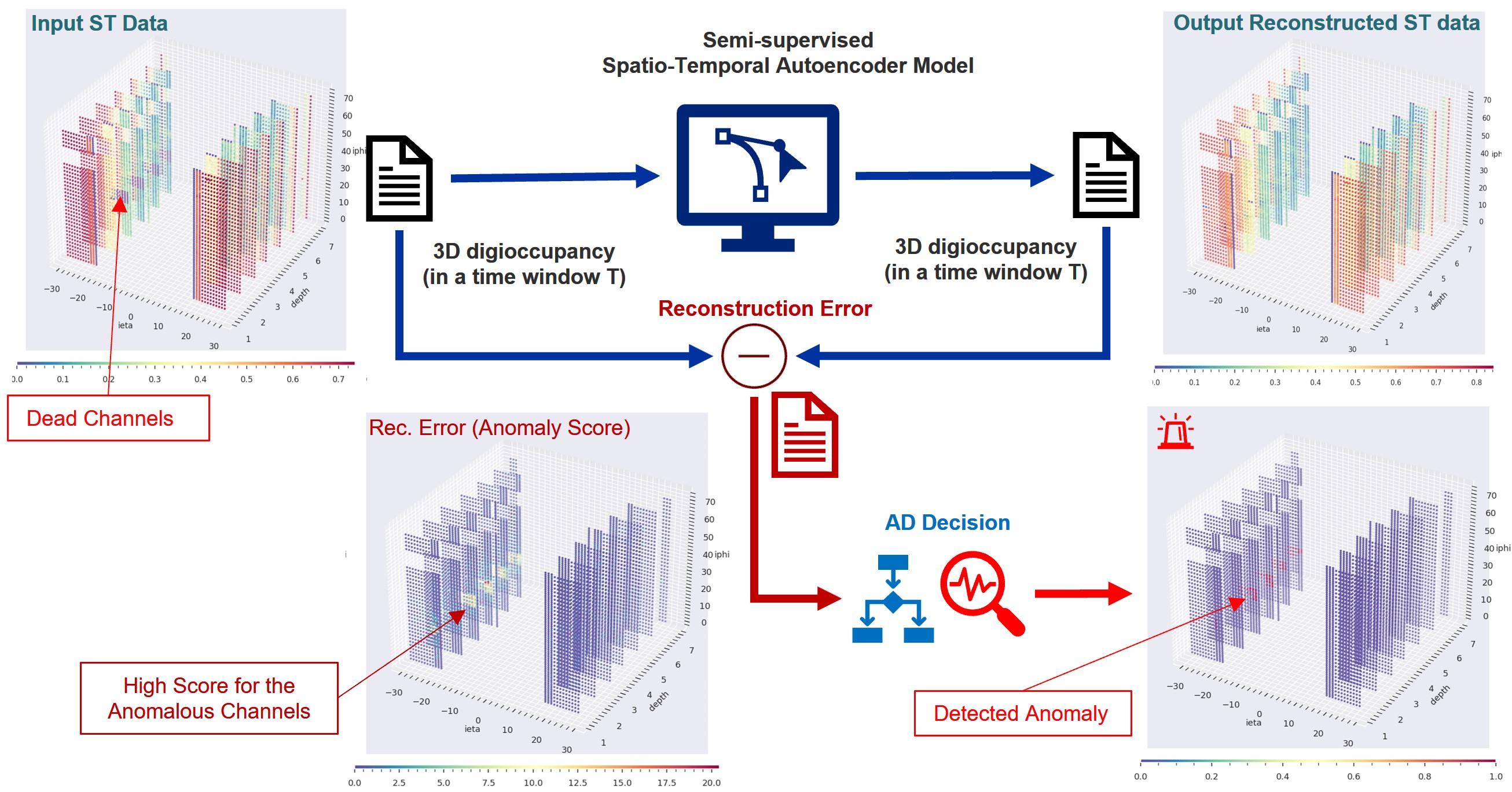}
\caption{The proposed channel-localized AE reconstruction AD system. The AE reconstructs the input ST digi-occupancy map, and a spatial AD decision is performed using the anomaly scores estimated from the ST reconstruction errors.}
\label{fig:ad_system_diagram}
\end{figure}

There is a lack of adequate labeled anomaly data covering all possible channel fault scenarios for the HCAL, and the anomalies may follow unpredictable patterns with different severity, shape, and size.  
We thus employed a semi-supervised approach for the AD system---{GraphSTAD} system; we trained a deep AE model to reconstruct healthy digi-occupancy maps with low contamination of anomalies. 
We present an ST reconstruction AE to detect abnormality in the HCAL channels using reconstruction deviation scores on ST digi-occupancy maps from consecutive lumisections (see Figure \ref{fig:ad_system_diagram}). 
The AE combines CNNs, GNNs, and RNNs to capture ST characteristics of digi-occupancy maps. 
The spatial feature extraction of the CNNs is leveraged with GNNs to learn circuit and housing-connectivity-induced spatial behavior irregularities among the HCAL sensor channels.
There are approximately {7000} channels---pixels---on the digi-occupancy map of the HCAL endcap subsystem, housed in 36 RBXes. The channels in a given RBX are susceptible to system faults in the RBX due to the shared backbone circuit and environmental factors like temperature and humidity. 
Our proposed GraphSTAD employs GNNs in its spatial feature extraction network pipeline to capture the characteristics of the HCAL channels owing to their shared physical connectivity to a given RBX.
GNNs have recently achieved promising results in several applications at the LHC~\cite{duarte2022graph, shlomi2020graph} and outperformed CNNs in learning irregular calorimeter geometry~\cite{qasim2019learning} and in pileup mitigation~\cite{martinez2019pileup}. 
The GraphSTAD system exploits both CNNs and GNNs~\cite{bruna2013spectral, kipf2016semi} to capture Euclidean and non-Euclidean spatial characteristics of the HCAL channels, respectively.

\subsection{Data Preprocessing}

This section explains the data preprocessing stages of our proposed AD approach: (1) digi-occupancy renormalization, and (2) graph adjacency matrix generation.

\subsubsection{Digi-Occupancy Map Renormalization}

\begin{figure}[H]
\includegraphics[width=0.9\columnwidth, scale=1]{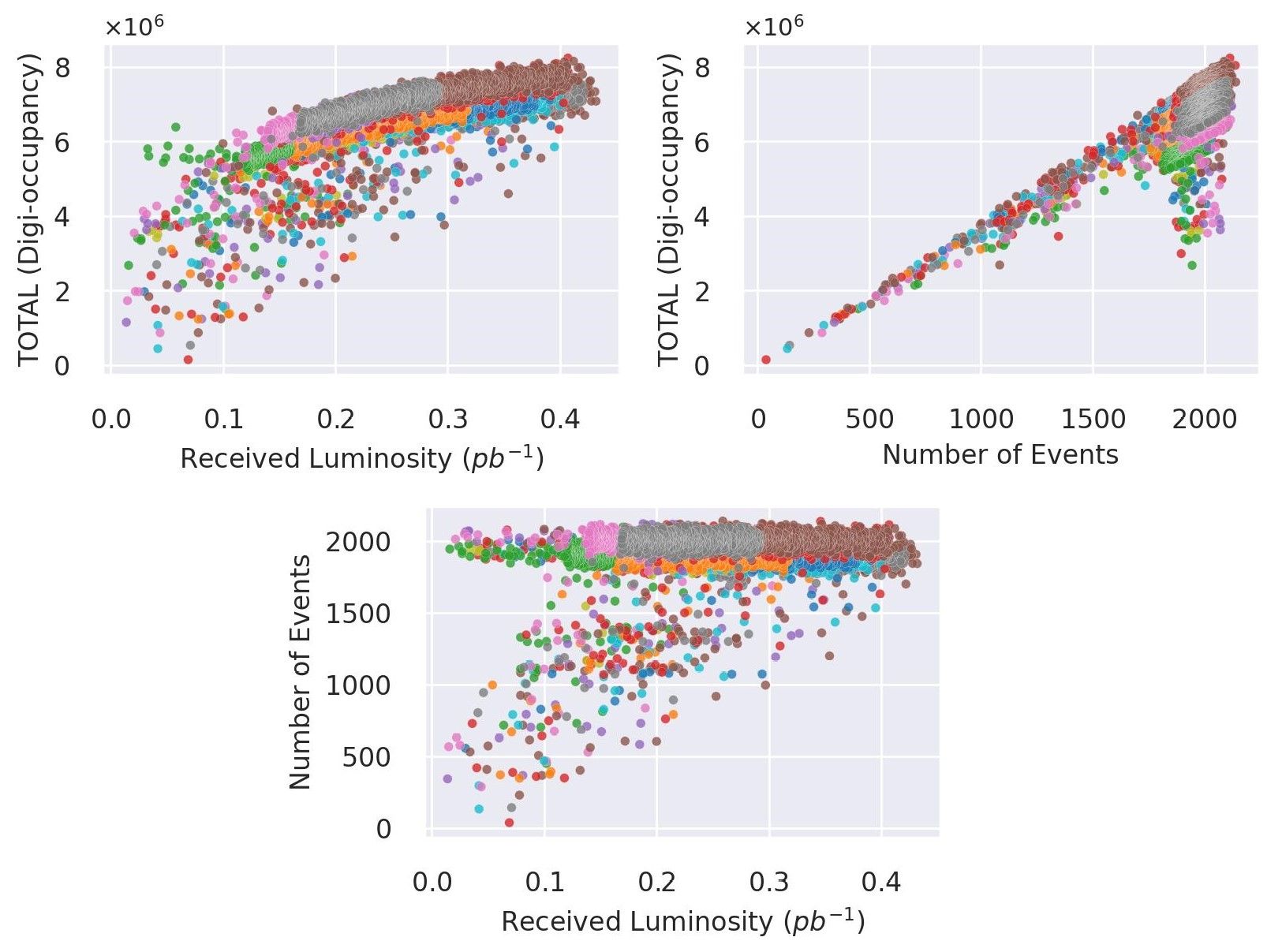}
\caption{Digi-occupancy and run settings---the received luminosity and the number of events---in LS granularity. The number of events did not fully follow the drop in luminosity (\textbf{bottom} plot) and digi-occupancy (\textbf{top-right} plot), {in contrast to the simultaneous shift in luminosity and digi-occupancy (\textbf{top-left} plot)}---portraying the nonlinear behavior of the LHC. The different colors correspond to different collision runs.
}
\label{fig:digioccuancy_RecLum_NumEvent_scatter}
\end{figure}

The digi-occupancy ($\gamma$) map data of the HCAL vary with the received luminosity ($\beta$) and the number of events ($\xi$) (see Figure \ref{fig:digioccuancy_RecLum_NumEvent_scatter}). 
We devised a renormalization of $\gamma$ through a regression model $\mathcal{R}$ to have a consistent quantity interpretation of $\gamma$ and build an AD model that robustly generalizes previously unseen run settings---$\beta$ and $\xi$ variations. 
The model $\mathcal{R}$ estimates the renormalizing $\bar{\gamma}_s$ at the $s$th LS using $\beta$ and $\xi$ as: 
\begin{equation}
\label{eq:digi_norm_reg}
\bar{\gamma} _{s} = \mathcal{R}(\xi, \beta).
\end{equation}

The model $\mathcal{R}$ is trained to minimize the MSE cost function,  $\mathbb{E}[(\gamma_s-\bar{\gamma}_{s})^2]$, where $\gamma_s$ is calculated as: 
\begin{equation}
\label{eq:digioccp_tot}
    \gamma_ s = \sum_{\forall i}^{}\gamma (s, i),
\end{equation}
where $\gamma (s, i)$ is the {digi-occupancy} of the $i${th} channel in the map at the $s${th} LS. 
Finally, the per-channel $\gamma (s, i)$ is renormalized by its corresponding $\bar{\gamma}_ s$ as:
\begin{equation}
\label{eq:digioccp_norm}
    \hat{\gamma} (s, i) = \frac{K\gamma (s, i)}{\bar{\gamma}_s},
\end{equation}
where $\hat{\gamma}$ is the renormalized $\gamma$, and $K$ is a scaling factor to compensate for the difference in the number of channels on the depth axes.

We employ fully connected ($\operatorname{FC}$) neural networks to build the regression model to effectively capture the nonlinear relationships illustrated in Figure \ref{fig:digioccuancy_RecLum_NumEvent_scatter}:
\begin{equation}
\label{eq:reg_model}
input(\xi, \beta) \operatorname{\to ReLU(FC(64)) \to ReLU(FC(64)) \to ReLU(FC(7)) \to} output(\bar{\gamma}_s). 
\end{equation}

\begin{figure}[H]
\includegraphics[width=0.7\columnwidth, scale=1]{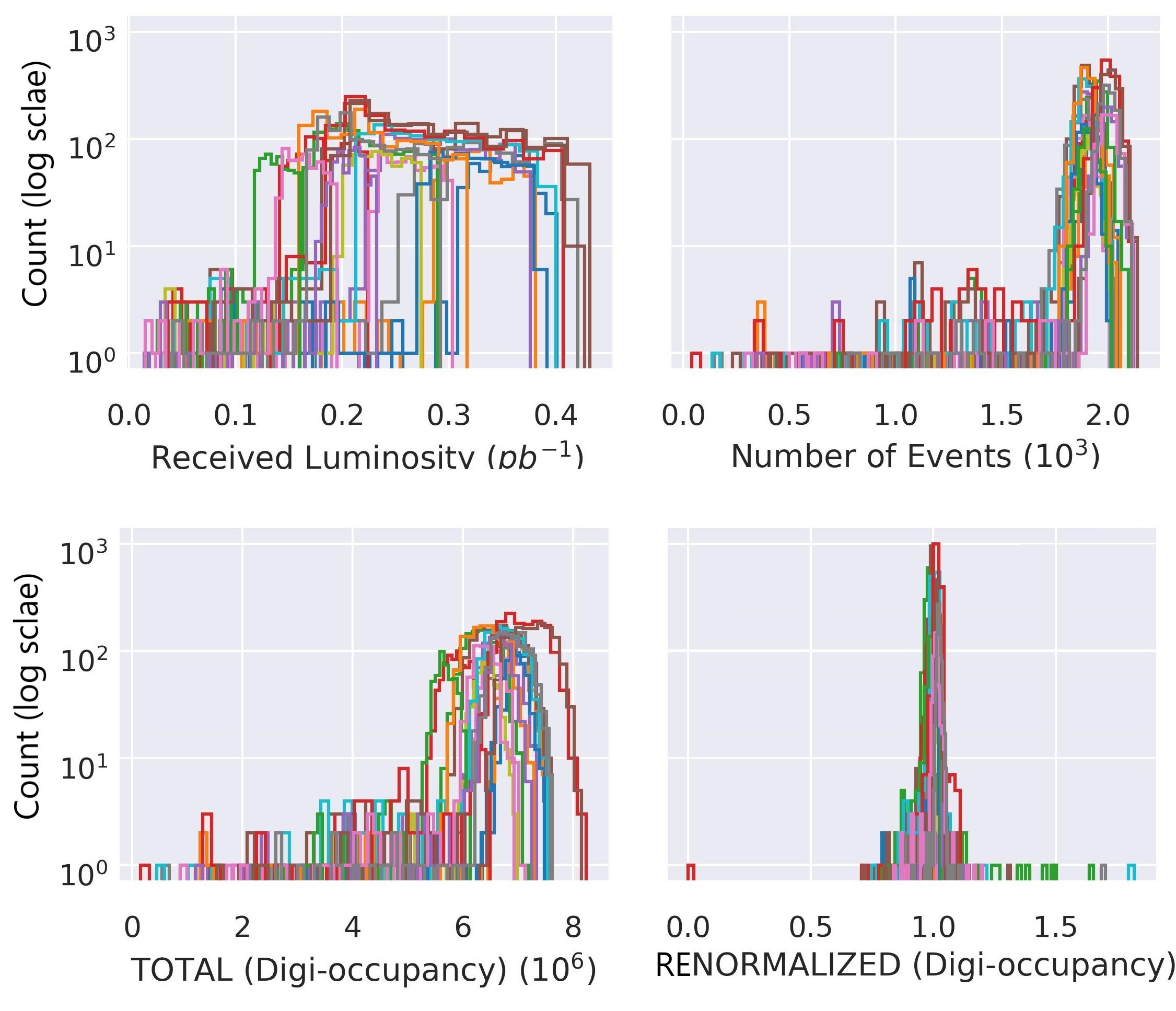}
\caption{Distribution of total {digi-occupancy} per LS before and after renormalization. From left to right: (\textbf{top}) the received luminosity, and the number of events; (\textbf{bottom}) the {digi-occupancy}, and the renormalized {digi-occupancy} obtained with the regression model described in the text. The different colors correspond to different runs.
}
\label{fig:digioccuancy_norm_distplot}
\end{figure}

Figure \ref{fig:digioccuancy_norm_distplot} depicts the data distribution of $\gamma_ s$ before and after renormalization with $\mathcal{R}$. The renormalization has successfully handled the discrepancies on the $\gamma_ s$ from several runs; it overlaps and centers distributions of $\hat{\gamma}_s$ and minimizes the outliers.  

\subsubsection{Adjacency Matrix Generation for Graph Network}
 
We employ an undirected graph network $\mathcal{G}(\mathcal{V}, \Theta)$ to represent the calorimeter channels in a graph network based on their connection to a shared RBX system. The graph $\mathcal{G}$ contained nodes $\upsilon \in \mathcal{V}$, with edges $ (\upsilon _i, \upsilon _j) \in \Theta$ in a binary adjacency matrix $\mathcal{A} \in \mathbb{R}^{M \times M}$, where $M$ is the number of nodes. An edge indicated the channels sharing the same RBX as:

\begin{equation}
\label{eq:adjacency_mtrix}
    A(\upsilon _i, \upsilon _j) = 
\begin{cases}
   1,& \text{if } \Omega (\upsilon _i) = \Omega (\upsilon _j) \\
    0,              & \text{otherwise},
\end{cases}
\end{equation}
where $\Omega (\upsilon)$ returns the RBX ID of the channel $\upsilon$.
There are about 7K channels in a graph representation of the digi-occupancy map of the HE calorimeter---each RBX network contains roughly 190 nodes. We retrieved the channel to RBX mapping from the calorimeter segmentation map of the HE.

\subsection{Anomaly Detection Modeling with Autoencoder Model}

We denote the AE model of the GraphSTAD system as $\mathcal{F}$. The ST data,\linebreak $\mathrm{X} \in \mathbb{R}^{T \times N_{i\eta} \times N_{i\phi} \times N_d \times N_f}$, are represented as a sequence in a time window $ t_x \in [t-T, t]$, where $N_{i\eta} \times N_{i\phi} \times N_d$ is the spatial dimension corresponding to the $i\eta$, $i\phi$, and $depth$ axes, respectively, and $N_f=1$ is the number of input variables---only a digi-occupancy quantity in the spatial data. 
The $\mathcal{F}_{\theta, \omega}: \mathrm{X} \to \bar{\mathrm{X}}$ is parameterized by $\theta$ and $\omega$ and attempts to reconstruct the input ST data $\mathrm{X}$ and outputs $\bar{\mathrm{X}}$.  
The encoder network of the model $E_\theta: \mathrm{X} \to \mathbf{z}$ provides the low-dimension latent space $\mathbf{z} = E_\theta(\mathrm{X})$, and the decoder $D_\omega: \mathbf{z} \to \bar{\mathrm{X}}$ reconstructs the ST data from $\mathbf{z}$, $\bar{\mathrm{X}}=D_\omega(\mathbf{z})$ as: 

\begin{equation}
\label{eq:ae}
\bar{\mathrm{X}} = \mathcal{F}_{\theta, \omega}(\mathrm{X}) = D_\omega (E_\theta(\mathrm{X})).
\end{equation}

The channel anomalies can be transients---short-lived and impacting only a single digi-occupancy map---or persist over time---affecting a sequence of maps. 
The spatial reconstruction error $e$ to detect a transient anomaly is calculated as:
\begin{equation}
\label{eq:rec_loss}
e_i = |x_i-\bar{x}_i|,
\end{equation}
where $x_i \in \mathrm{X}$ and $\bar{x}_i \in \bar{\mathrm{X}}$ are the input and reconstructed {digi-occupancy} of the $i${th} channel. $e_i$ detects a channel abnormality occurrence on isolated maps. 
We opted for an aggregated error in a time window $T$ using the mean absolute error (MAE) to capture a time-persistent anomaly as:
\begin{equation}
\label{eq:mae_rec_loss}
e_{i, MAE} = \frac{1}{T} \sum^t_{t^\prime=t-T}e_i(t^\prime).
\end{equation}

We standardized $e_i$ to regularize the reconstruction accuracy variations among the channels, allowing a single AD decision threshold $\alpha$ to all the channels in the spatial map:
\begin{equation}
\label{eq:anomaly_score}
s_i = \frac{e_i}{\sigma _i},
\end{equation}
where $\sigma _i$ is the standard deviation of $e_i$, or $e_{i, MAE}$ if the time window is considered, on the training data set. 
The anomaly flags $a_i$ are generated after applying $\alpha$ to the anomaly scores, $a_i = s_i > \alpha$.
$\alpha$ is a tunable constant that controls the detection sensitivity.

\subsection{Autoencoder Model Architecture}

\begin{figure}[H]
\includegraphics[width=1\columnwidth, scale=1]{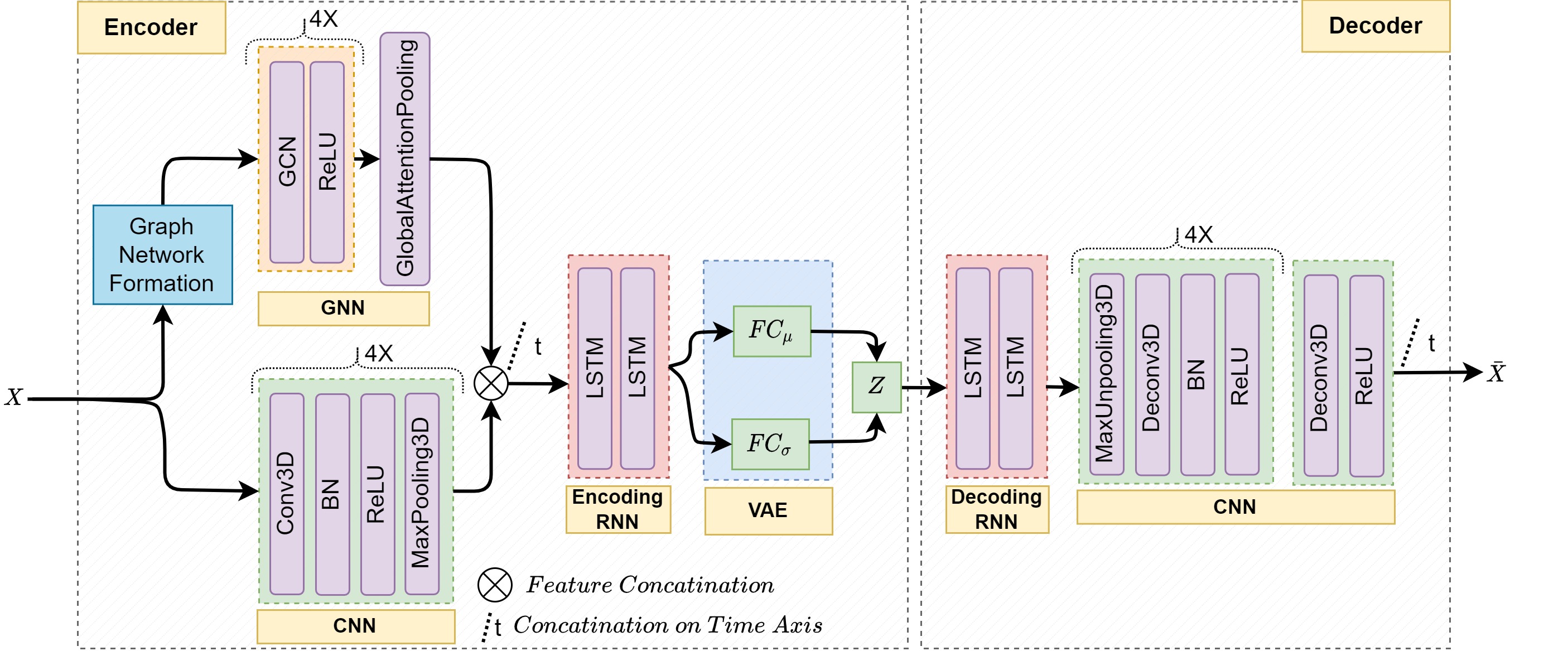}
\footnotesize{\par
Conv3D: 3D convolutional neural network; GCN: graph convolutional neural networks; Deconv3D: 3D deconvolutional neural networks; BN: batch normalization; LSTM: long short-term memory recurrent networks; FC: fully connected neural networks.
\par }
\caption{The architecture of the proposed AE for the GraphSTAD system. The GNN and CNN are for spatial feature extraction at each time step, and the RNN captures the temporal behavior of the extracted features. The encoder incorporates the GNN for backend physical connectivity among the spatial channels, CNN for regional spatial proximity of the channels, and RNN for temporal behavior extraction. The decoder contains RNN and deconvolutional neural networks to reconstruct the spatio-temporal input data from the low-dimensional latent features. 
}
\label{fig:propose_autoencoder_detailed_diagram}
\end{figure}

Convolutional neural networks have achieved state-of-the-art performance in several applications, mainly with image data~\cite{chang2022video, luo2019video, hasan2016learning, hsu2017anomaly, wu2020fast}. The shared nature of the kernel filters of CNNs substantially reduces the number of trainable parameters in the model compared to fully connected neural networks.
Directly supplying the learned spatial features to temporal neural networks, such as RNNs, could become inherently challenging for high-dimensional data due to computational overhead. 
We employed CNNs and GNNs with a pooling mechanism to extract relevant features from high-dimensional spatial data, followed by RNNs to capture temporal characteristics of the extracted features (see Figure \ref{fig:propose_autoencoder_detailed_diagram}).
We integrated the variational layer~\cite{kingma2013auto} at the end of the encoder for overfitting regularization by enforcing continuous and normally distributed latent representations~\cite{asres2021unsupervised, guo2018multidimensional, an2015variational, chadha2019comparison}.

The CNN of the encoder has $L_c$ networks, each containing $\operatorname{Conv3D} (\cdot, kernel\_size=[3 \times 3 \times 3])$ for regular spatial learning followed by batch normalization ($\operatorname{BN}$) for network weight regularization and faster convergence, $\operatorname{ReLU}$ for nonlinear activation, and $\operatorname{MaxPooling3D}$ for spatial dimension reduction. The model can be summarized as: 
\begin{equation}
\label{eq:encoder_cn_arch_1}
y^c_t, \psi ^c_t = \operatorname{Pool}(\operatorname{ReLU}(\operatorname{BN}(\operatorname{Conv3D}(x_t^{l}, N_c^l)))) |_{l=1,\dots, L_c},
\end{equation}
where $x_t^l$ is the input spatial $\gamma$ map data at time step $t$, and $N_c^l$ is the feature size of the $l$th network. $y^c_t$ is the extracted feature set of the CNN at $t$. $\operatorname{Pool}(\cdot)$ denotes $\operatorname{MaxPooling3D}(\cdot, stride=[2 \times 2 \times 2])$. $\psi ^c_t$ holds the pooling spatial location indices of the $\operatorname{MaxPooling3D}$ layers to be used later for upsampling in the decoder during map reconstruction. 
The final extracted feature set $\mathrm{Y}_c \in \mathbb{R}^{T \times N_c}$ of the CNN is an aggregation of all $y^c_t$ in the time window $T$, concatenated on the time dimension:    
\begin{equation}
\label{eq:encoder_cn_arch_2}
\mathrm{Y}_c= [y^c_1, y^c_2, ..., y^c_T].
\end{equation}

We used $L_c = 4$ to map the input spatial dimension $[64 \times 72 \times 7]$ into $[4 \times 4 \times 1]$, which yielded a reduction factor of $2^{L_c}$ and expanded the feature space of the input from $N_f=1$ to $N_c=128$. $N^\prime_c: [4\times4\times1\times128]=2048$ features were generated after reshaping. 

The GNN of the encoder has $L_g$ networks of a graph convolutional network {($\operatorname{GCN}$)} with a $\operatorname{ReLU}$ activation, and a final {global attention pooling}~\cite{wang2019dgl}. The networks are summarized as: 

\begin{equation}
\label{eq:encoder_gn_arch_1}
\begin{gathered}
y^g_t = \operatorname{Pool}(\operatorname{ReLU}(\operatorname{GCN}(x_t^{l}, N_g^l) |_{l=1,\dots, L_g})), \\
\mathrm{Y}_g = [y^g_1, y^g_2, ..., y^g_T],
\end{gathered}
\end{equation}
where the $\operatorname{GCN}$ layers have a feature size of $N_g^l$, and $\operatorname{Pool}(\cdot)$ signifies the\linebreak $\operatorname{GlobalAttentionPooling}(\cdot)$ at the end of the GNN. $\operatorname{GlobalAttentionPooling}$ aggregates the graph node features with an attention mechanism to obtain the final feature set of the GNN $\mathrm{Y}_g \in \mathbb{R}^{T \times N_g}$.   
Similar to the CNN, we set $L_g = 4$ and $N_g = 128$ to generate the $\mathrm{Y}_g$. 

The encoded ST feature set $\zeta \in \mathbb{R}^{1 \times N_z}$ is obtained by learning the temporal context on the extracted spatial features $\mathrm{Y}=[\mathrm{Y}_c, \mathrm{Y}_g]$ with two layers of long short-term memory ($\operatorname{LSTM}$) as:
\begin{equation}
\label{eq:encoder_rnn_arch}
\zeta = \operatorname{LSTM} (\mathrm{Y}, N^l_r)|_{l=1,2},
\end{equation}
where $N^l_r$ is the feature size of the $l$th LSTM layer. The last layer ($N^2_r = N_z=32$) generates the low-dimensional latent representation of the encoder.
The VAE layer of the encoder generates the normally distributed representation latent features $\mathbf{z}$ as:
\begin{equation}
\label{eq:encoder_vae_arch}
\begin{gathered}
\mathbf{z} =\mu _z+ \sigma _z\odot \epsilon,
\end{gathered}
\end{equation}
where $\odot$ signifies an element-wise product with a standard normal distribution sampling $\epsilon \sim \mathcal{N}(0, 1)$~\cite{an2015variational}. $\mu _z$ and $\sigma _z$ of the VAE are implemented with $\operatorname{FC}$ layers taking $\zeta$ as input. 

The decoder network of the AE comprises an RNN and a CNN to reconstruct the target ST data from the latent features. The decoding starts with a temporal feature reconstruction using an $\operatorname{LSTM}$ network as: 
\begin{equation}
\label{eq:decoder_rnn_arch}
\bar{\zeta} = \operatorname{LSTM} (\mathbf{z}, N_r^l)|_{l=1,2},
\end{equation}
where $\bar{\zeta}$ is the reconstructed temporal feature set from the latent space $\mathbf{z}$.
A spatial reconstruction follows for each time step $t$ through a multilayer deconvolutional neural network. Each network starts with $\operatorname{MaxUnpooling3D}(\cdot, stride=[2 \times 2 \times 2], \psi^c_l)$ to upsample the spatial data using localization indices $\psi^c_l$ from the $l$th $\operatorname{MaxPooling3D}$ of the encoder followed by a deconvolutional layer ($\operatorname{Deconv3D} (\cdot, kernel\_size=[3 \times 3 \times 3])$)~\cite{zeiler2010deconvolutional}, a $\operatorname{BN}$, and a $\operatorname{ReLU}$. Eventually, $\operatorname{Deconv3D} (\cdot, kernel\_size=[1 \times 1 \times 1])$ is incorporated for the final output stabilization. 
The decoder network is summarized as: 
\begin{equation}
\label{eq:decoder_cn_arch_1}
\begin{gathered}
\bar{x}_t = \operatorname{ReLU}(\operatorname{BN}(\operatorname{Deconv3D}(\operatorname{Unpool}(\bar{\zeta_t}, \psi^c_t), N_c^l)) |_{l=1,\dots, L_c} \\
\bar{x}_t = \operatorname{ReLU}(\operatorname{Deconv3D}(\bar{x}_t, N_f)),
\end{gathered}
\end{equation}
where $\bar{x}_t$ is the reconstructed spatial data, and $\operatorname{Unpool}(\cdot)$ denotes $\operatorname{MaxUnpool3D}(\cdot)$.
The final reconstructed ST data $\bar{\mathrm{X}} \in \mathbb{R}^{T \times N_{i\eta} \times N_{i\phi} \times N_d \times N_f}$ are obtained as:    
\begin{equation}
\label{eq:decoder_cn_arch_2}
\bar{\mathrm{X}}= [\bar{x}_1, \bar{x}_2, ..., \bar{x}_T].
\end{equation}

\subsection{Model Training}
We trained the AE on healthy digi-occupancy maps of LHC collision runs.
The modeling task became a multivariate learning problem since the target data contained readings from multiple calorimeter channels in the spatial digi-occupancy map. 
An appropriate scaling of the spatial data was thus necessary for effective model training; we further normalized the spatial data per channel into a range of $[0, 1]$. 
We also observed that the $\gamma$ distribution of the channels at the first depth of the spatial map was different from the channels at the higher depths (see Figure \ref{fig:he_digioccupancy_sample}); a distribution imbalance on target channel data affects model training efficacy when well-known statistical algorithms, e.g., MSE, are employed as loss functions. The MSE loss minimizes the cost of the entire space, and it may converge to a nonoptimal local minimum in the presence of an imbalanced data distribution; this phenomenon is known as the class imbalance challenge in machine learning classification problems. A widely used remedy is to employ a weighting mechanism---assigning weights to the different targets. We applied a weighted MSE loss function to scale the loss from the different distributions, the $depth \in {1}$ and $depth \in {2,\dots,7}$:
\begin{equation}
\label{eq:weighted_training_loss}
\mathcal{L}^{\prime} = \sum_{j}\frac{\varsigma _j}{M_j} \sum_{i \in C_j}(x_i -\bar{x}_i)^2,
\end{equation}
where $x_i$ is the $\hat{\gamma}$ of the $i${th} channel in the $j${th} group set $C_j$, $M_j$ is the number of channels in $C_j$, and $\varsigma _j$ is the weight factor of the MSE loss of the $j${th} group.
We holistically set $\varsigma_1 = 0.4$ and $\varsigma_2=1$ after experimenting with different $\varsigma$ values. 

The VAE regularized the training MSE loss using the KL divergence loss $D_{KL}$ to achieve the normally distributed latent space as: 
\begin{equation}
\label{eq:training_loss}
\mathcal{L} = \underset{W \in \mathbb{R}} {\text{argmin}}
\left\{ \mathcal{L}^{\prime} - \beta {D_{KL}}\left[ \mathcal{N}(\mu_z, \sigma_z) \| \mathcal{N}(0, I)\right] + \rho \|W\|_{2}^{2} \right\},
\end{equation}
where $\mathcal{N}$ is a normal distribution with zero mean and unit variance, and $\|.\|$ is the \textit{Frobenius norm} of the $L_2$ regularization for the trainable model parameters $W$. $\beta=0.003$ and $\rho=10^{-7}$ are tunable regularization hyperparameters.
We finally used the \textit{Adam} optimizer with superconvergence via \textit{one-cycle} learning rate scheduling~\cite{smith2019super} for training.

\section{Experimental Results and Discussion}
\label{sec:resultsanddiscussion}

\begin{figure}[H]
\includegraphics[width=0.7\textwidth, scale=1]{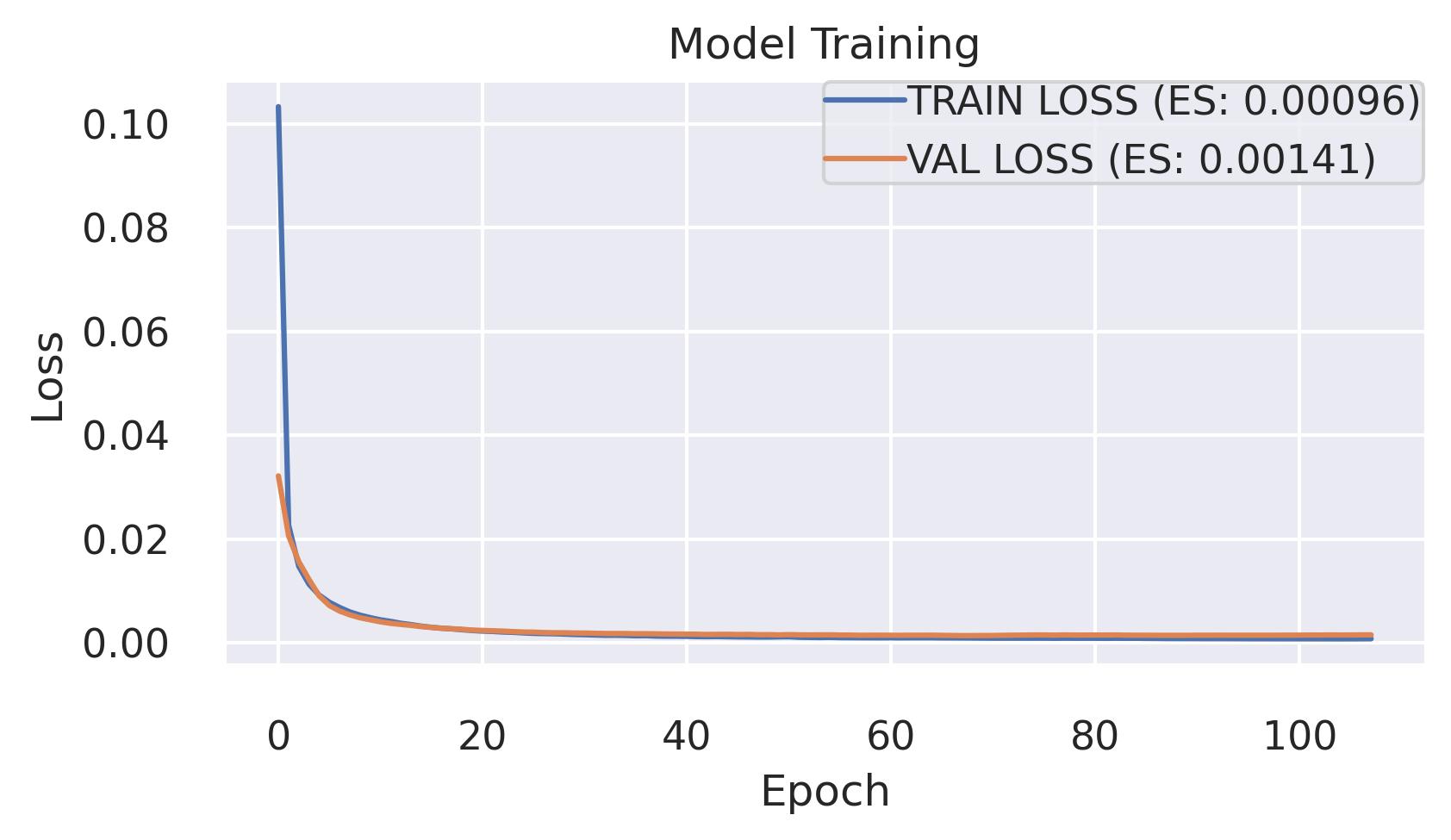}
\caption{GraphSTAD autoencoder model training (\textit{early stopping = $20$ epochs, learning rate = $10^{-3}$, weight regularization = $10^{-7}$, training time = $82$} min). The low training loss indicates a good model fitting---no underfitting---to the data set, and the low validation loss demonstrates a good generalization---no overfitting.}
\label{fig:model_training_loss}
\end{figure}

AD studies for the DQM inject simulated anomalies into good data to validate the effectiveness of the developed models since a small fraction of the data is affected by real anomalies~\cite{azzolin2019improving}. 
Likewise, we trained the GraphSTAD autoencoder model using four GPUs on {10,000} digi-occupancy maps---from LS sequence number $[1, 500]$---and evaluated on LSs $[500, 1500]$ injected with synthetic anomalies simulating real dead, hot, and degraded calorimeter channels. We employed early stopping using 20\% of the training dataset to estimate the validation loss during each training epoch (see Figure \ref{fig:model_training_loss}). 
The model training achieved good fitting and generalization, as demonstrated by the low loss and closeness between the training and validation losses.

\begin{figure}[H]
\includegraphics[width=0.9\textwidth, scale=1]{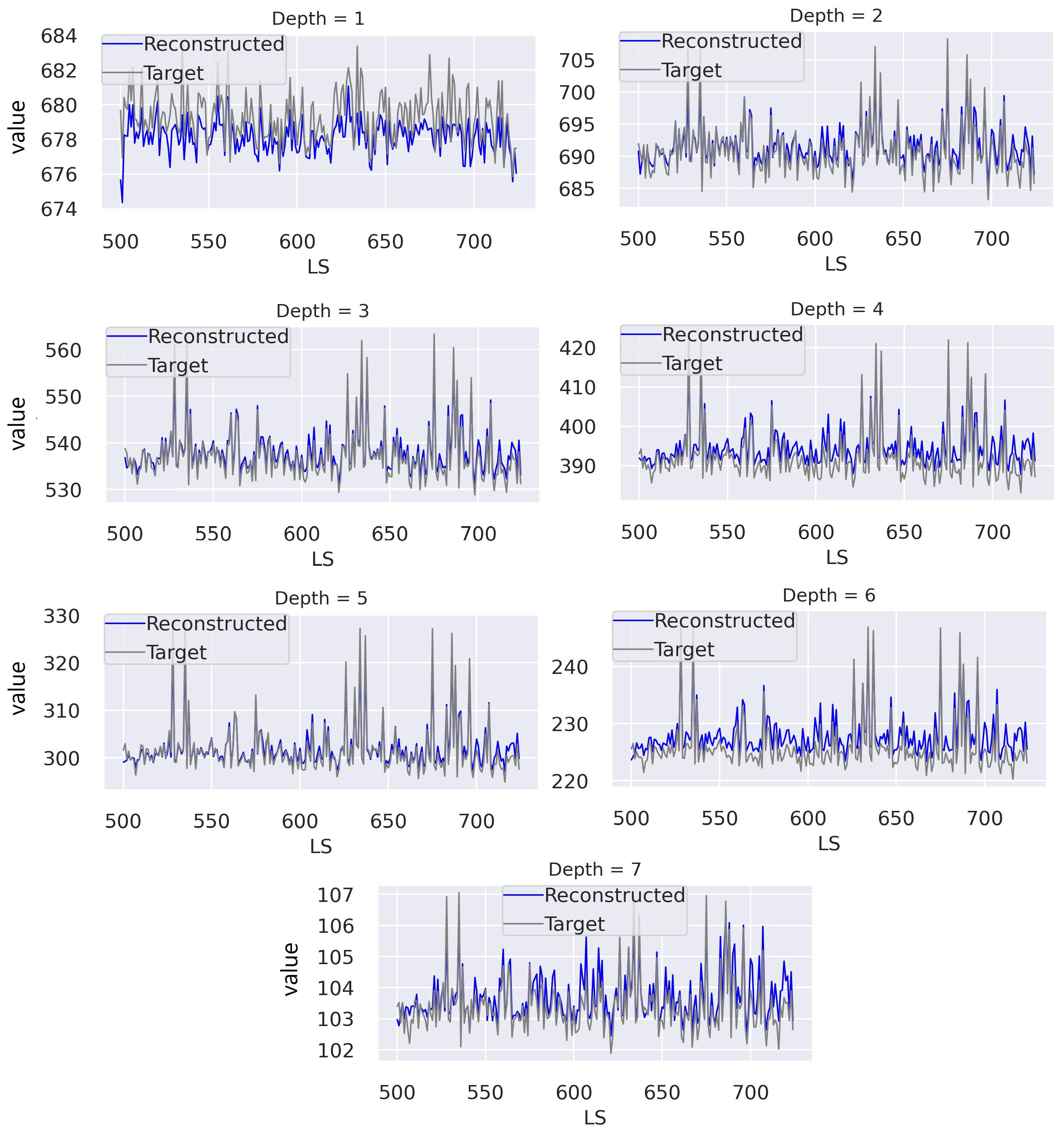}
\caption{ST digi-occupancy maps' reconstruction on samples from the test data set (\textit{RunId: 325170, LS = [500, 750]}). The figure illustrates the total {digi-occupancy} across the seven depths---$\hat{\gamma}_ l$. Our GraphSTAD AE operates on ST $\gamma$ map data, and we present the above plots, corresponding to the $\gamma_ l$ per LS, to demonstrate the capability of the AE in handling the fluctuation across the sequence of LSs.}
\label{fig:total_dgioccupancy_recon_lineplot}
\end{figure}

Figure \ref{fig:total_dgioccupancy_recon_lineplot} demonstrates the capability of the proposed ST AE in reconstructing normal digi-occupancy maps from a sequence of lumisections. The AE accomplished a promising reconstruction ability on the ST digi-occupancy data. A high reconstruction accuracy on the healthy data is essential to reduce false-positive flags when a semi-supervised AE is employed for AD application. 
We further discuss the reconstruction error distribution comparison on the healthy and abnormal channels in Section \ref{sec:eval_degrading_channel}.

We discuss below the performance of our proposed model, comparisons with benchmark models, detection results on real faulty channels, and model complexity cost.

\subsection{Anomaly Detection Performance}
\label{sec:perfwithcounters}

We created synthetic anomalies to simulate dead, hot, and degraded channels and then injected them into healthy digi-occupancy maps. We subsequently evaluated the ability of the AD to detect the injected anomalies. 
The anomaly generation algorithm involved three steps: (1) a selection of a random set of LSs $\tau \in[500, 1500]$ from the test set, (2) a random selection of spatial locations $\varphi$ for each $\tau$, where $\varphi\in \{i\eta  \times i\phi  \times depth\}$ on the HE axes (see Figure \ref{fig:he_digioccupancy_sample}), and (3) injection of anomalies. 
The anomalies were simulated using degrading factor $R_D$ with $\gamma_{a} = R_D\gamma_{h}$, where $\gamma_{h}$ and $\gamma_{a}$ are the healthy and anomaly channel $\gamma$ values, respectively. The anomalies are dead ($R_D=0$, and $\gamma_{a} = 0$), hot ($R_D>1$, and $\gamma_a >> \gamma_h$), and degraded ($0 < R_D \in < 1$, and $0 \leq \gamma_a < \gamma_h$) channels.
We kept the same $\tau$ and $\varphi$ as for the generated anomalies for evaluation consistency on the different anomaly types.

\subsubsection{Detection of Dead and Hot Channels}

We evaluated the AD accuracy on dead- ($\gamma_a = 0, ~R_D=0$) and hot- ($\gamma_a = R_D\gamma_h, ~R_D=200\%$) channels on the {10,000} maps---{5000} maps for each anomaly type. Tables \ref{tbl:clf_per_channel_dead_hot} and \ref{tbl:clf_per_channel_dead_hot_tw} present the AD performance on transient anomalies---short-lived in isolated maps---and time-persisting anomalies---encroaching consecutive maps in a time window---respectively. Our model achieved a good accuracy with precise localization of the faulty channels, precision of 0.99 when capturing 99\% of the faulty channels. 
Time-persistent anomalies were easier to detect; the FPR generally improved by 13--23\% and 28--40\% for the dead and hot anomalies, respectively, compared to the short-lived anomalies on isolated LSs. We observed that most false positives (FPs) occurred on channels with a low expected $\gamma_h$, where the model achieved a relatively lower reconstruction accuracy. The performance was not entirely unexpected since we trained the AE to minimize a global MSE loss function \eqref{eq:training_loss}. The reconstruction errors became relatively high for channels with low $\gamma$ ranges that limited the effectiveness in distinguishing the anomalies when capturing 99\% of the time-persistent dead channels using \eqref{eq:mae_rec_loss}.

We monitored roughly 31.28 {million} HE sensor channels, of which {335,000} (1.07\%) were simulated abnormal channels, from the {5000} maps on the isolated map evaluation in Table \ref{tbl:clf_per_channel_dead_hot}. The monitored channels grew to 156 {million} with 1.68 {million} (1.07\%) anomalies for the evaluation of time-persistent anomalies in Table \ref{tbl:clf_per_channel_dead_hot_tw} using five time-window maps resulting in {25,000} maps. 

\begin{table}[ht]
\caption{AD on dead and hot channel anomalies on isolated digi-occupancy maps.\label{tbl:clf_per_channel_dead_hot}}
\newcolumntype{C}{>{\centering\arraybackslash}X}
\begin{tabularx}{\textwidth}{CCCCCC}
\toprule
\textbf{Anomaly Type}                  & \textbf{Captured Anomalies} & \textbf{P} & \textbf{R} & \textbf{F1} & \textbf{FPR} \\
\midrule
\multirow{3}{*}{Dead Channel} & 99\%               & $0.999$    & $0.99$     & $0.995$     & $6.722 \times 10^{-6}$ \\ 
                                      & 95\%               & $1.000$    & $0.95$     & $0.974$     & $3.102 \times 10^{-6}$ \\ 
                                      & 90\%               & $1.000$    & $0.90$     & $0.947$     & $2.068 \times 10^{-6}$ \\ \hline
\multirow{3}{*}{Hot Channel}  & 99\%               & $0.999$    & $0.99$     & $0.994$     & $9.113 \times 10^{-6}$ \\ 
                                      & 95\%               & $1.000$    & $0.95$     & $0.974$     & $1.939 \times 10^{-6}$ \\ 
                                      & 90\%               & $1.000$    & $0.90$     & $0.947$     & $1.196 \times 10^{-6}$ \\
\bottomrule
\end{tabularx}
\noindent{\footnotesize{P---precision, R---recall, F1---F1-score, FPR---false positive rate}}
\end{table}
\vspace*{-1\baselineskip}

\begin{table}[htbp]
\caption{AD on time-persistent dead and hot channel anomalies.\label{tbl:clf_per_channel_dead_hot_tw}}
\newcolumntype{C}{>{\centering\arraybackslash}X}
\begin{tabularx}{\textwidth}{CCCCCC}
\toprule
\textbf{Anomaly Type}                  & \textbf{Captured Anomalies} & \textbf{P} & \textbf{R} & \textbf{F1} & \textbf{FPR} \\
\midrule
\multirow{3}{*}{Dead Channel} & 99\%               & $0.999$    & $0.99$     & $0.995$     & $7.691\times 10^{-6}$ \\  
                                      & 95\%               & $1.000$    & $0.95$     & $0.974$     & $2.715\times 10^{-6}$  \\  
                                      & 90\%               & $1.000$    & $0.90$     & $0.947$     & $1.616\times 10^{-6}$ \\ \hline
\multirow{3}{*}{Hot Channel}  & 99\%               & $0.999$    & $0.99$     & $0.995$     & $5.461\times 10^{-6}$  \\  
                                      & 95\%               & $1.000$    & $0.95$     & $0.974$     & $1.357\times 10^{-6}$ \\  
                                      & 90\%               & $1.000$    & $0.90$     & $0.947$     & $7.756\times 10^{-7}$  \\
\bottomrule
\end{tabularx}
\end{table}
\vspace*{-1\baselineskip}

\subsubsection{Detection of Degrading Channels}
\label{sec:eval_degrading_channel}

Table \ref{tbl:clf_per_channel_degrading_tw} presents the AD accuracy of time-persistent degraded channels simulated with $R_D = [80\%, 60\%, 40\%, 20\%, 0\%]$; $R_D=0\%$ corresponds to a dead channel. We injected the generated channel faults into {1000} maps for each decay factor. We monitored around 156 {million} channels, of which 1.74 {million} (1.11\%) were abnormal channels, from the total of {25,000} digi-occupancy maps---{5000} maps per time window.
The AD system demonstrated a promising potential in detecting degraded channel anomalies. The FPR to capture 99\% of the anomaly was 2.988\%, 0.155\%, 0.022\%, 0.002\%, and 0.001\% when channels operated at 80\%, 60\%, 40\%, 20\%, and 0\% of their expected capacity, respectively. 

\begin{figure}[H]
\includegraphics[width=1\textwidth, scale=1]{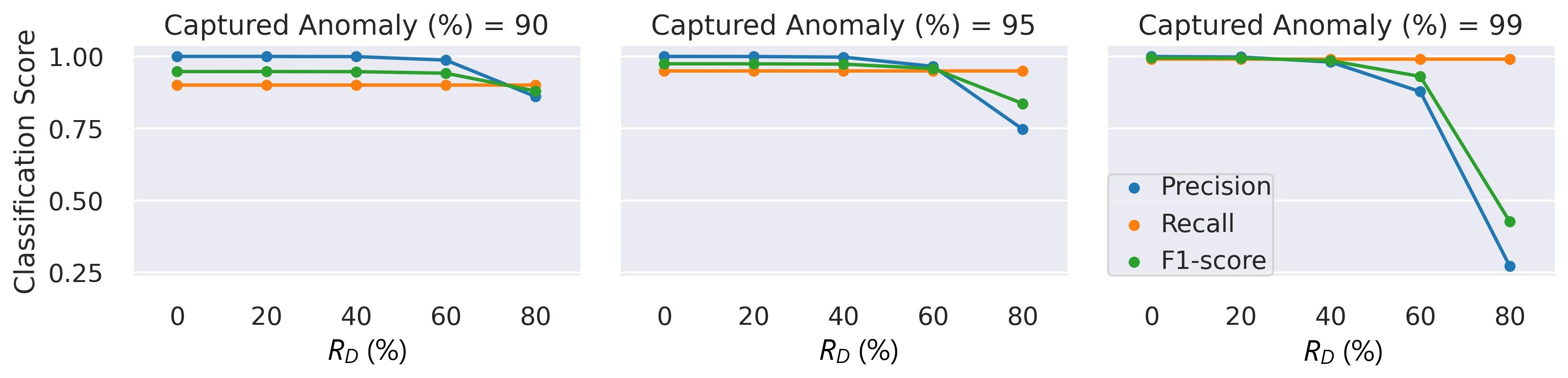}
\caption{AD classification performance on time-persistent degraded channels.}
\label{fig:clf_perf__window_spatial_scaled}
\vspace*{-1\baselineskip}
\end{figure}

\begin{figure}[H]
\includegraphics[width=0.75\textwidth, scale=1]{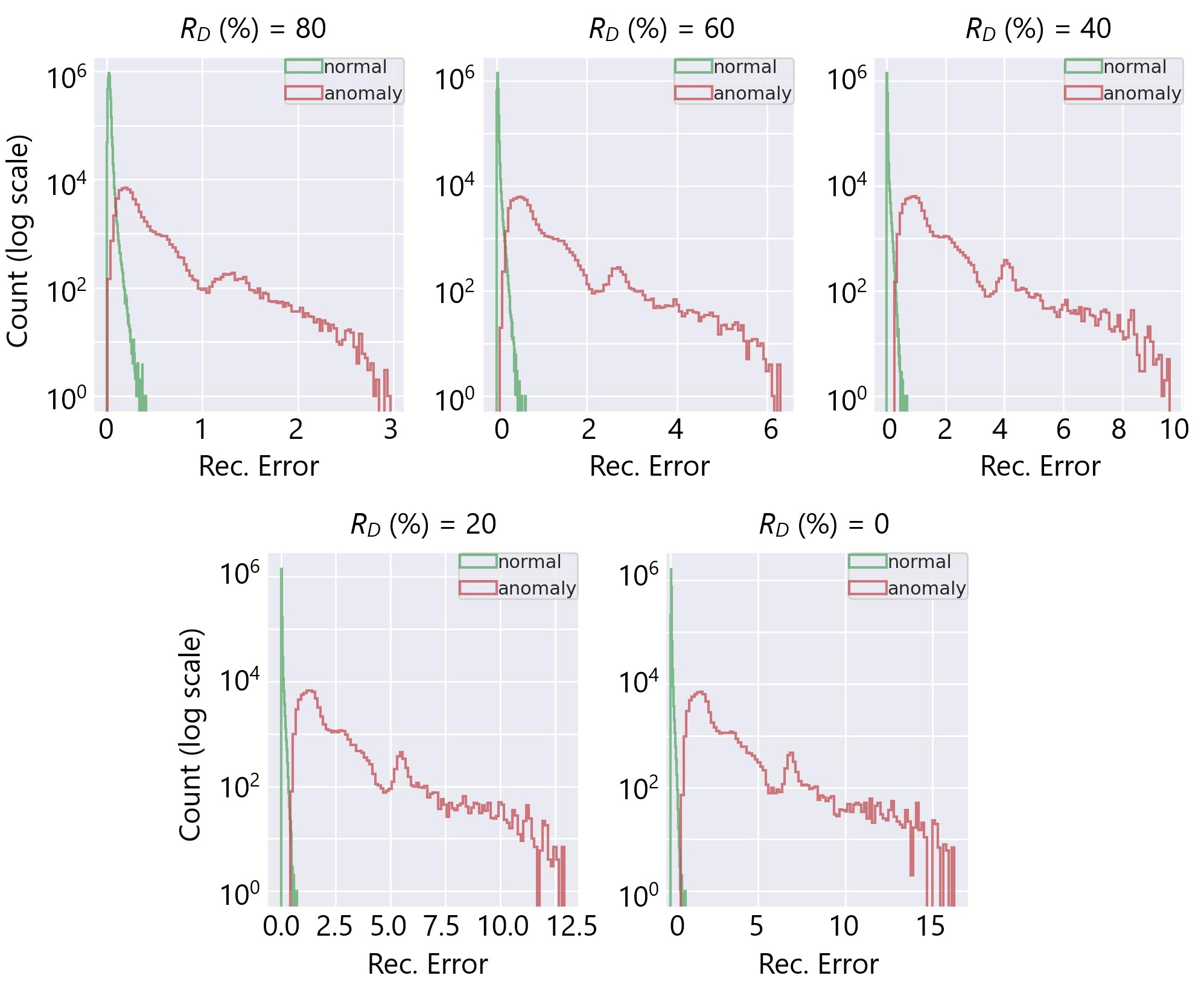}
\caption{Reconstruction error distribution of healthy and anomalous channels at different $R_D$'s. The overlap region decreases substantially as the channel deterioration increases (left to right).}
\label{fig:degrade_hot_to_death_channel__rec_err__window_spatial}
\end{figure}

The relatively lower precision at $R_D=80\%$ indicated that there were still a few anomalies challenging to catch despite the very low FPR considering the accurate classification of numerous true-negative healthy channels (see Figure \ref{fig:clf_perf__window_spatial_scaled}); the channels operating at $R_D=80\%$ were mostly inliers overlapping with the healthy operating ranges, and detecting them was difficult when the expected $\gamma_h$ of the channel was low. 
The significant improvement of the FPR by 88\% and 95\% when the number of captured anomalies was reduced to 95\% and 90\%, respectively, demonstrated that a small percentage of the channels caused the performance drop at $R_D=80\%$. 
Figure \ref{fig:degrade_hot_to_death_channel__rec_err__window_spatial} illustrates the overlap regions on the distribution of the reconstruction errors of the healthy and faulty channels at the various $R_D$ values. 

\begin{table}[htbp]
\caption{AD on time-persistent degraded channels. \label{tbl:clf_per_channel_degrading_tw}}
\newcolumntype{C}{>{\centering\arraybackslash}X}
\begin{tabularx}{\textwidth}{CCCCC}
\toprule
\textbf{Anomaly Type}                   & \boldmath{$R_D$}        & \multicolumn{1}{c}{\textbf{FPR (90\%)}} & \multicolumn{1}{c}{\textbf{FPR (95\%)}} & \multicolumn{1}{c}{\textbf{FPR (99\%)}} \\ 
\midrule
\multirow{5}{*}{\textbf{Degraded Channel}}    & 80\%                         & $1.636 \times 10^{-3}$                       & $3.614 \times 10^{-3}$                       & $2.988 \times 10^{-2}$                       \\  
                                          & 60\%                         & $1.329 \times 10^{-4}$                       & $3.834 \times 10^{-4}$                       & $1.550 \times 10^{-3}$                       \\  
                                          & 40\%                         & $8.405 \times 10^{-6}$                       & $2.764 \times 10^{-5}$                       & $2.242 \times 10^{-4}$                       \\  
                                          & 20\%                         & $2.263 \times 10^{-6}$                       & $5.173 \times 10^{-6}$                       & $2.505 \times 10^{-5}$                       \\  
                                          & 0\%                          & $9.699 \times 10^{-7}$                       & $1.778 \times 10^{-6}$                       & $6.142 \times 10^{-6}$                       \\
\bottomrule
\end{tabularx}
\end{table}
\vspace*{-1\baselineskip}

\subsection{Performance Comparison with Benchmark Models}

\begin{figure}[H]
\includegraphics[width=0.9\textwidth, scale=1]{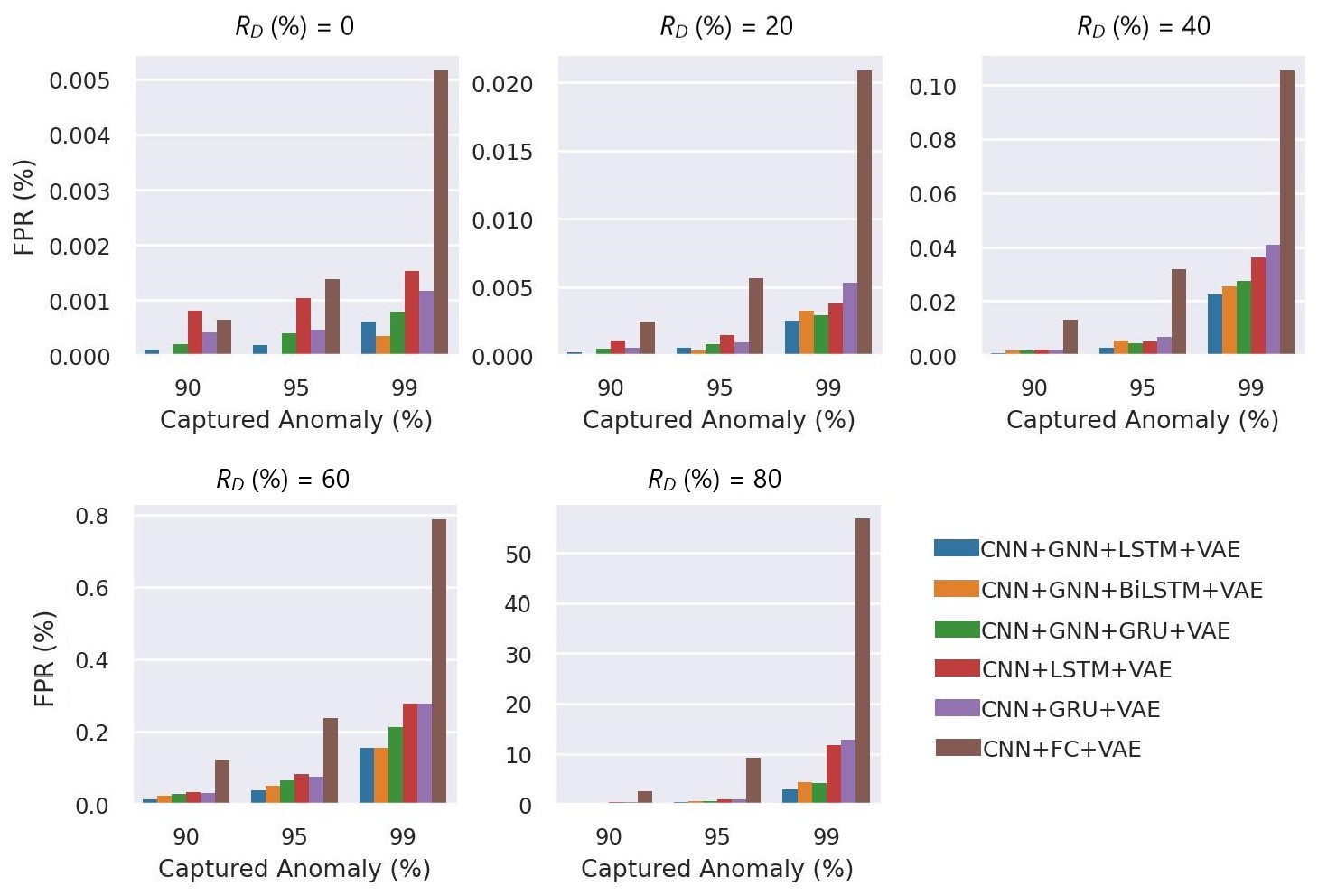}
\footnotesize{\par
CNNs: convolutional neural networks, GNNs: graph neural networks, BiLSTM: bidirectional LSTM, GRU: gated recurrent unit, and VAE: variational AE.
\par}
\caption{Comparison with benchmark models on time-persistent anomaly channels. 
The GraphSTAD (CNN + GNN + LSTM + VAE) achieved a significantly lower FPR.}
\label{fig:degrade_hot_to_death_channel_fpr_benchmark_compare}
\end{figure}

We quantitatively compared alternative benchmark models to validate the capability of GraphSTAD (see Figure \ref{fig:degrade_hot_to_death_channel_fpr_benchmark_compare}).
The benchmark AE models employed a similar architecture as the GraphSTAD AE but with different layers. The results demonstrated that the integration of the GNN had a significant performance improvement from 1.6 to 3.9 times in the FPR. The temporal models---with RNN---achieved a three- to fivefold boost over the nontemporal spatial AD model when capturing severely degraded channels. The GraphSTAD system had a substantial 25-time amelioration over the nontemporal model for subtle and inlier anomalies, e.g., channels deteriorated by 20\% at $R_D=80\%$. Incorporating temporal modeling and a GNN enhances the degraded channel detection performance.

\subsection{Detection of Real Anomalies in the HCAL}
\begin{figure}[H]
\captionsetup[subfigure]{justification=centering}
\begin{subfigure}[]{1\textwidth}
\centering
\includegraphics[width=1\textwidth, scale=1]{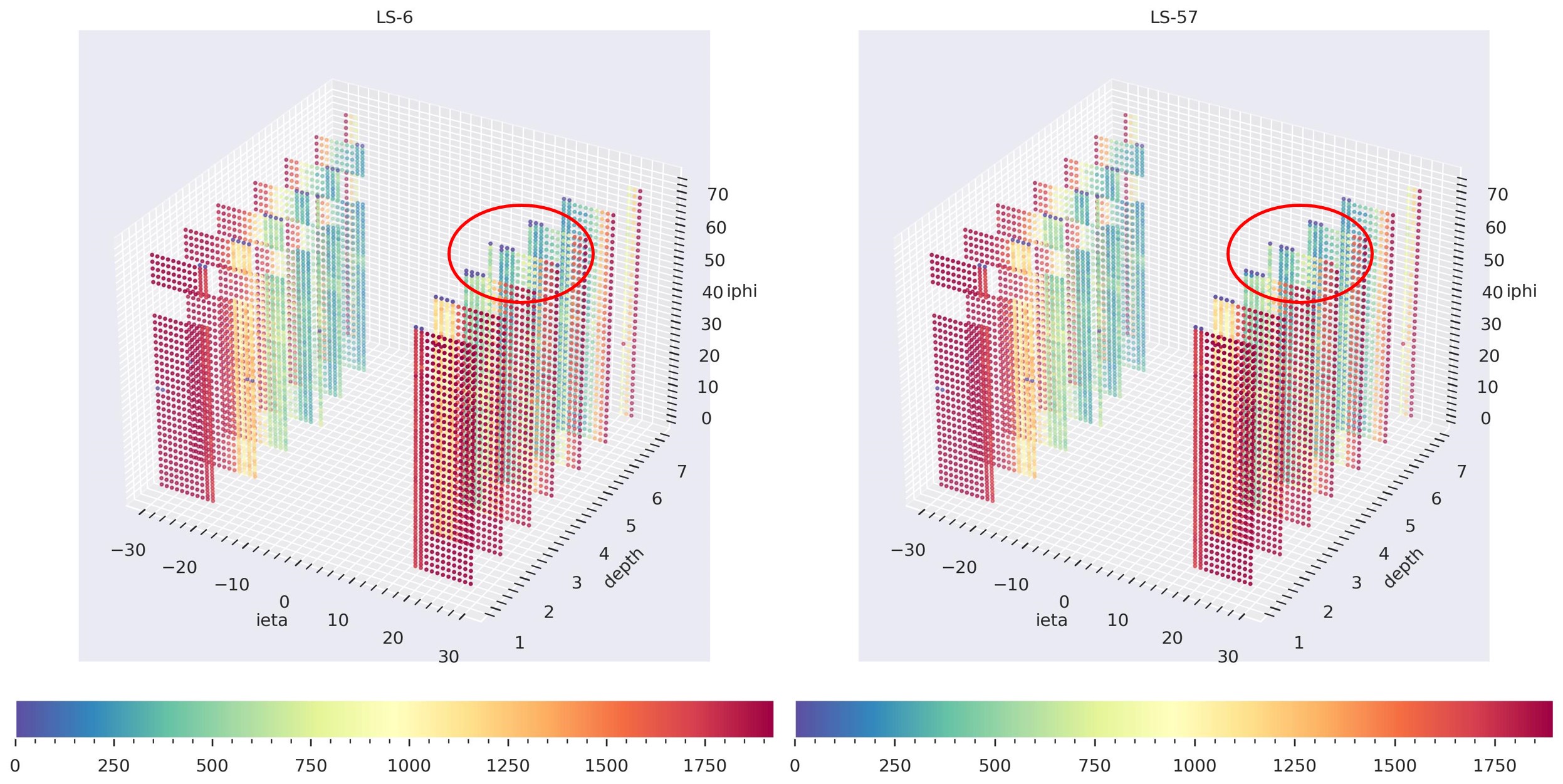}
\caption{}
 \medskip
\end{subfigure}

\begin{subfigure}[]{1\linewidth}
\centering
\includegraphics[width=1\textwidth, scale=1]{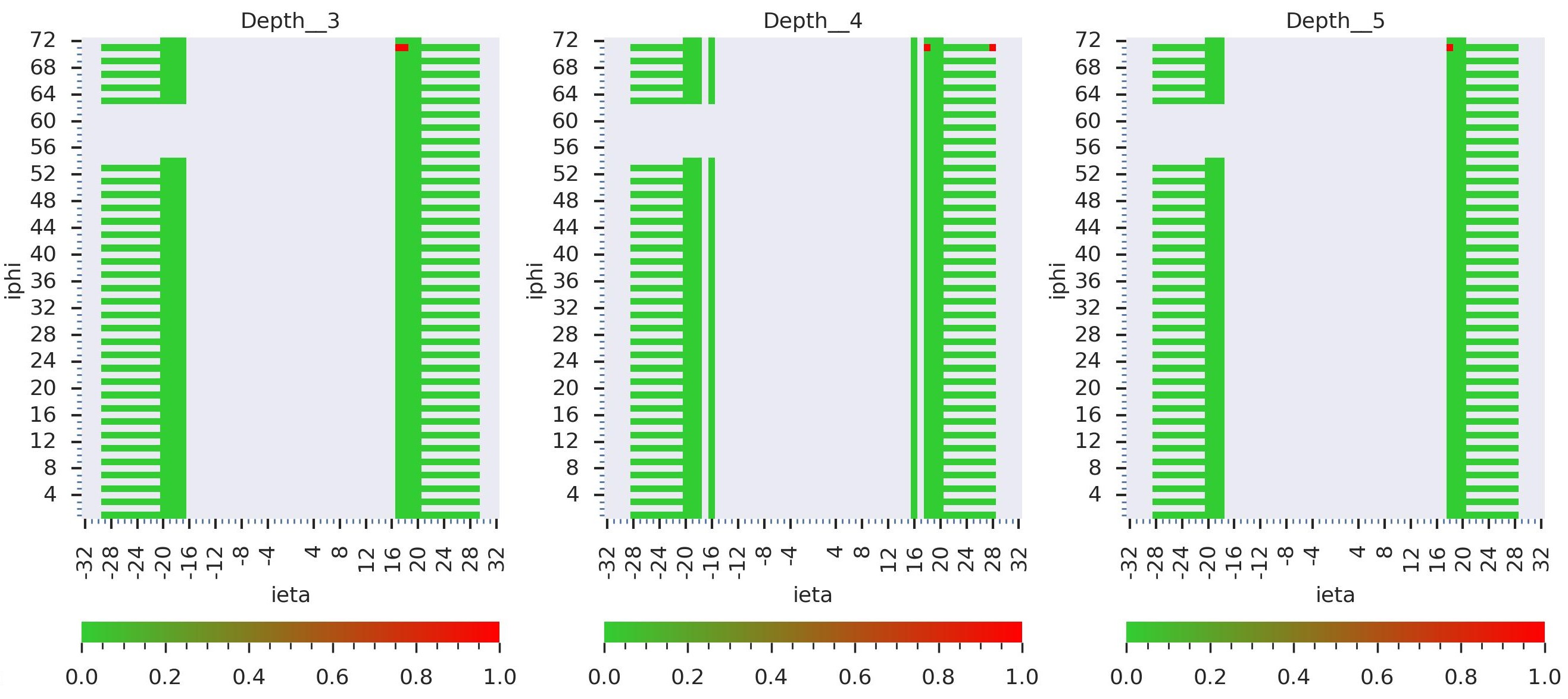}
\caption{}
\end{subfigure}
\caption{Spatial view on real faulty channels detection from \textit{RunId = 324841} collision run data. (\textbf{a}) The 3D digi-occupancy maps with faulty channels, dead on the left at \textit{LS = 6} and degraded on the right at \textit{LS = 57}, and (\textbf{b}) the anomaly flags on the 2D map according to the depth axes, red for an anomaly and green for healthy. Previously known bad channels during model training were excluded in the plots and were not detected as new.}
\label{fig:digioccupancy_anomaly_hist3d}
\end{figure}

\begin{figure}[H]
\captionsetup[subfigure]{justification=centering}
\begin{subfigure}[]{.95\linewidth}
\centering
\includegraphics[width=\textwidth, scale=1]{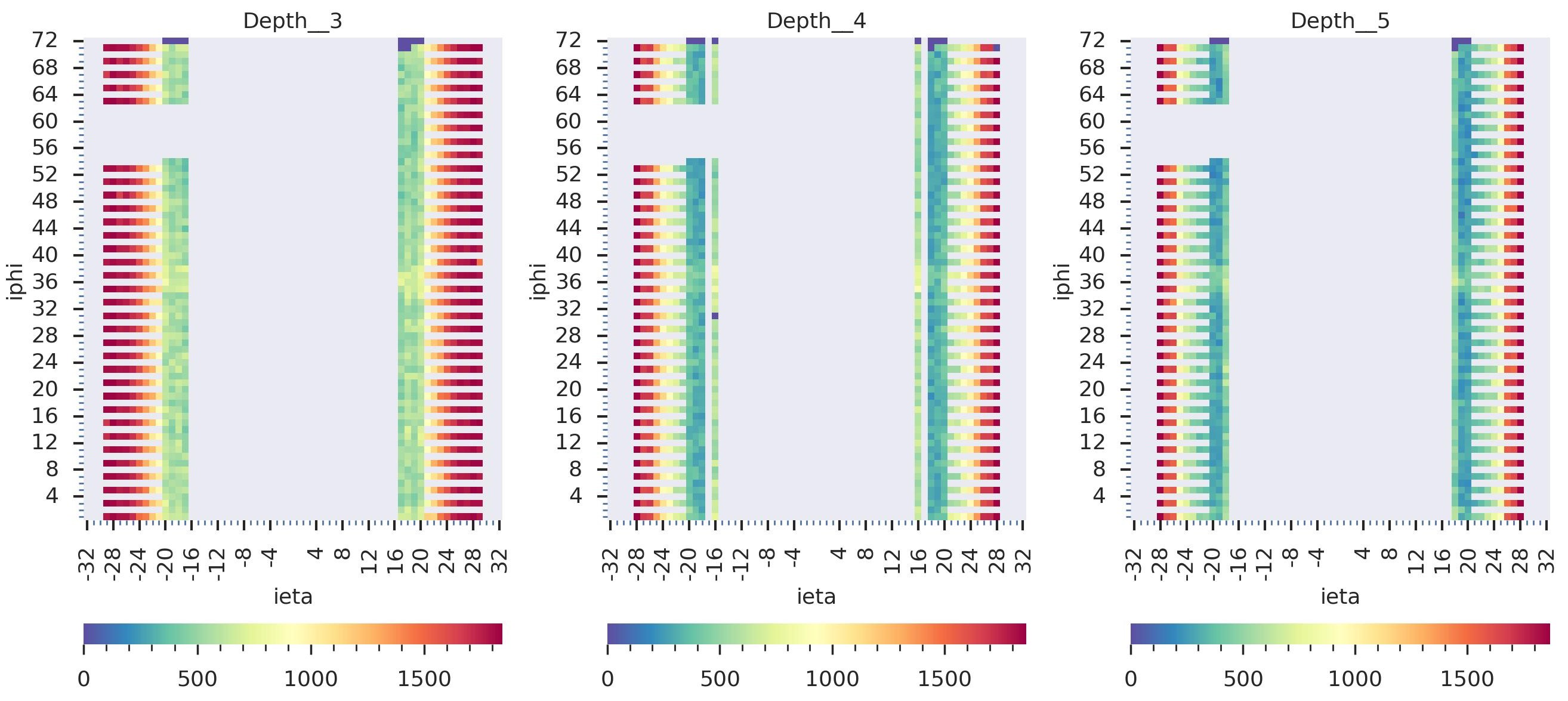}
\caption{}
 \medskip
\end{subfigure}
\begin{subfigure}[]{.95\linewidth}
\centering
\includegraphics[width=\textwidth, scale=1]{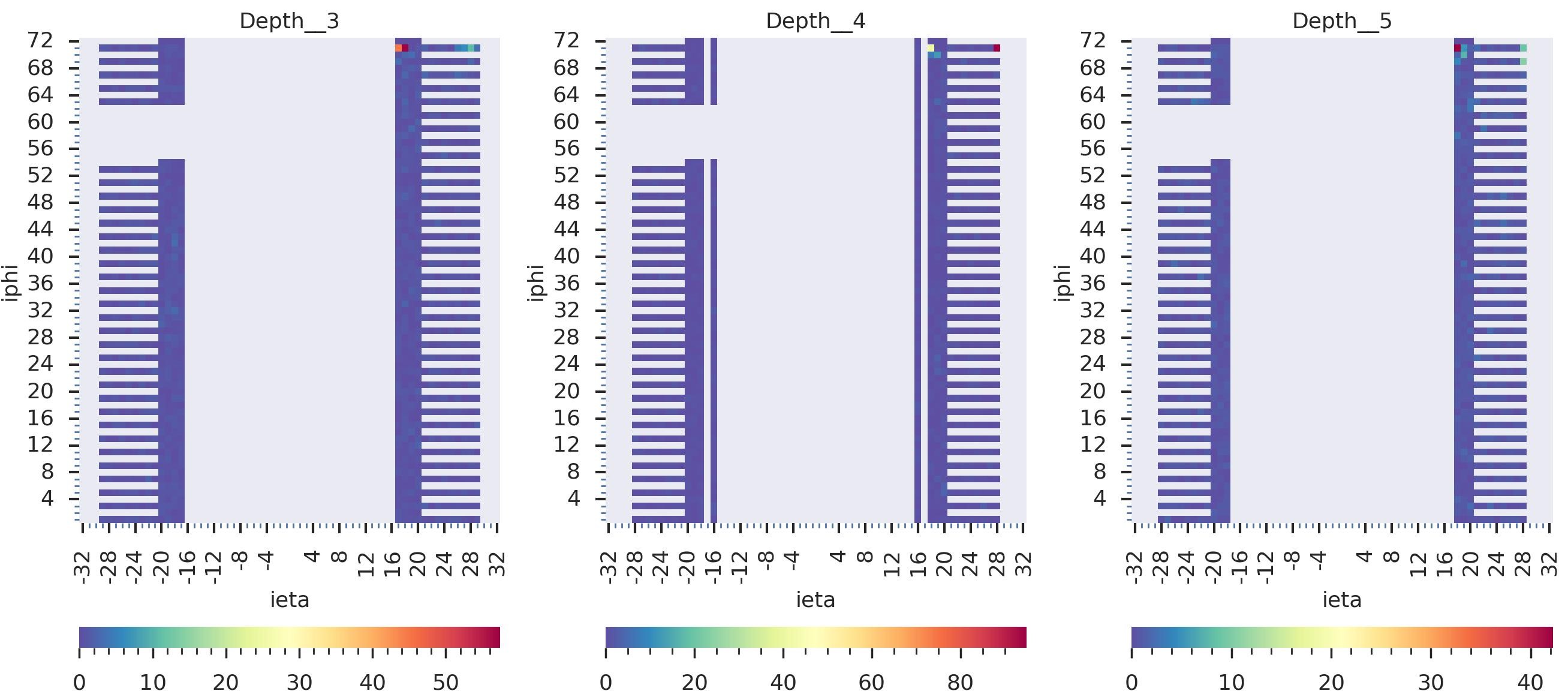}
\caption{}
\end{subfigure}
\caption{Spatial view on the detected real dead channels at \textit{LS = 6} from \textit{RunId = 324841}. (\textbf{a}) The raw 2D digi-occupancy maps at the \textit{depth} axes of the faulty channels and (\textbf{b}) the corresponding anomaly score maps. The GraphSTAD localized the anomaly scores on the faulty dead channels.
}
\label{fig:digioccupancy_anomaly_hist2d_ls10}
\end{figure}

Our GraphSTAD system caught five real faulty HE channels in collision data\linebreak \textit{RunId = 324841} using the digi-occupancy maps. 
The faulty channels were located at\linebreak $[i\eta, i\phi, depth]: [17, 71, 3]$, $[18, 71, 3]$, $[18, 71, 4]$, $[18, 71, 5]$, and $[28, 71, 4]$ and impacted 52 consecutive LSs (see Figure \ref{fig:digioccupancy_anomaly_score__line}). Figures \ref{fig:digioccupancy_anomaly_score__line} and \ref{fig:digioccupancy_anomaly_hist3d} illustrate the detected faults fell into the dead channel category except the last one \textit{LS = 57}, where the channels operated in a degraded state---the $\gamma$ was lower than expected. Detecting degraded channels is challenging since the $\gamma$ reading is nonextreme as in dead and hot channels, and the $\gamma$ drop overlaps with other false down-spikes (see \textit{LS $>$ 57} in Figure \ref{fig:digioccupancy_anomaly_score__line}). The down-spikes in the {digi-occupancy} for \textit{LS $>$ 57} are due to a nonlinearity in the LHC, changes in collision run settings (see Figure \ref{fig:digioccupancy_anomaly_score__line_setting}). Our normalizing regression model successfully handled the fluctuation during prepossessing before causing false-positive alerts (see Figure \ref{fig:digioccupancy_anomaly_score__digi_anml_score}). 
Figures \ref{fig:digioccupancy_anomaly_hist2d_ls10} and \ref{fig:digioccupancy_anomaly_hist2d_ls57} portray the spatial anomaly scores during the death and degraded status of the faulty channels;
the high scores localized at the faulty channels demonstrated the GraphSTAD AD performance at a channel-level granularity.
The existing production DQM system of the CMS uses rule-based and statistical methods and has also reported these abnormal channels in a run-level analysis; the results are only available at the end of the run after analyzing all the LSs for the run~\cite{tuura2010cms}. 
Our approach is adaptive to variability in the digi-occupancy maps and provides an anomaly localization that detects faulty channels, including nonextreme degraded channels, per lumisection granularity.

\begin{figure}[H]
\captionsetup[subfigure]{justification=centering}
\begin{subfigure}[]{1\textwidth}
\centering
\includegraphics[width=\textwidth, scale=1]{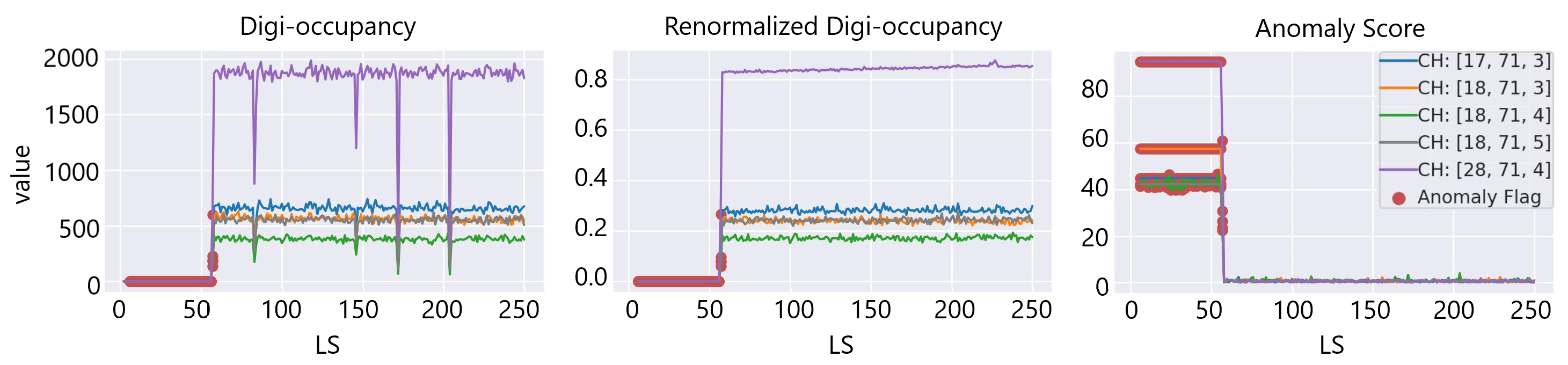}
\caption{}
\medskip
\label{fig:digioccupancy_anomaly_score__digi_anml_score}
\end{subfigure}
\caption{\textit{Cont}.}
\end{figure}

\begin{figure}[H]\ContinuedFloat
\captionsetup[subfigure]{justification=centering}
\begin{subfigure}[]{1\textwidth}
\centering
\includegraphics[width=\textwidth, scale=1]{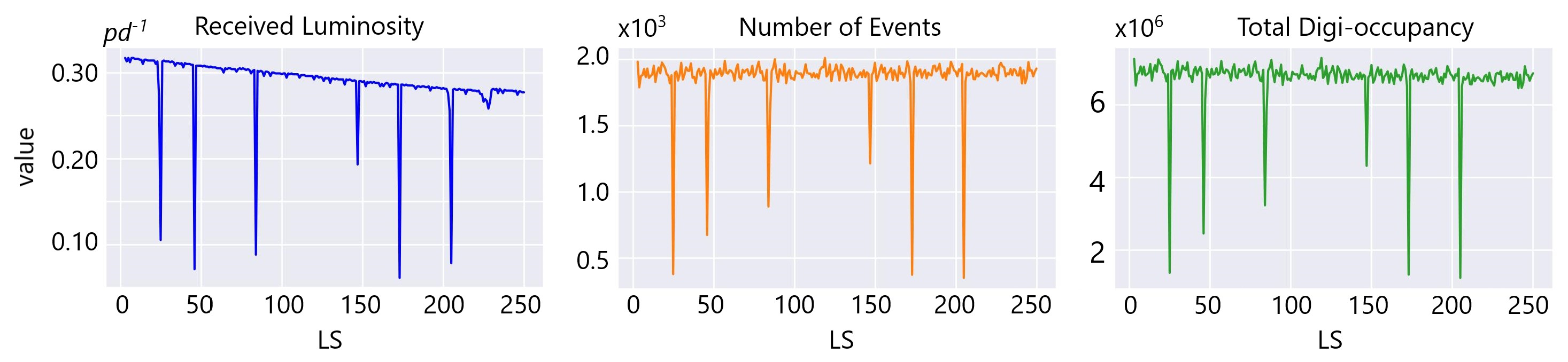}
\caption{}
\medskip
\label{fig:digioccupancy_anomaly_score__line_setting}
\end{subfigure}
\caption{Detected real faulty channels on digi-occupancy maps at \textit{LS = [6, 57]} of \textit{RunId = 324841}. (\textbf{a}) The {digi-occupancy} dropped to near zero for the faulty channels (left and middle plots), resulting in high anomaly scores (right). Dead (\textit{LS = [6, 56]}) and degraded channel anomalies (\textit{LS = 57}) were captured on the highlighted LSs (red). (\textbf{b}) Collision run settings and the total {digi-occupancy} per LS.}
\label{fig:digioccupancy_anomaly_score__line}
\end{figure}

\begin{figure}[H]
\captionsetup[subfigure]{justification=centering}
\begin{subfigure}[]{1\linewidth}
\centering
\includegraphics[width=\textwidth, scale=1]{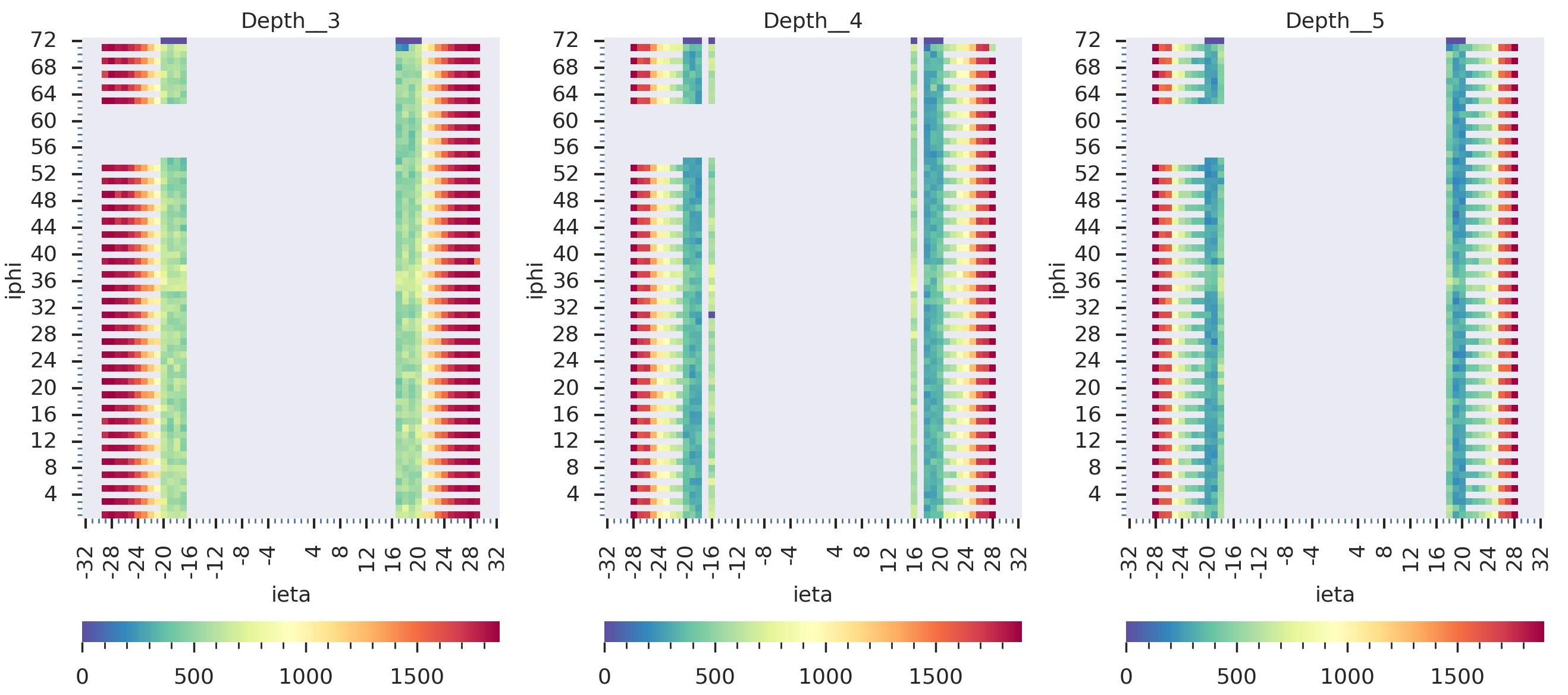}
\caption{}
 \medskip
\end{subfigure}
\caption{\textit{Cont}.}
\end{figure}

\begin{figure}[H]\ContinuedFloat
\captionsetup[subfigure]{justification=centering}
\begin{subfigure}[]{1\linewidth}
\centering
\includegraphics[width=\textwidth, scale=1]{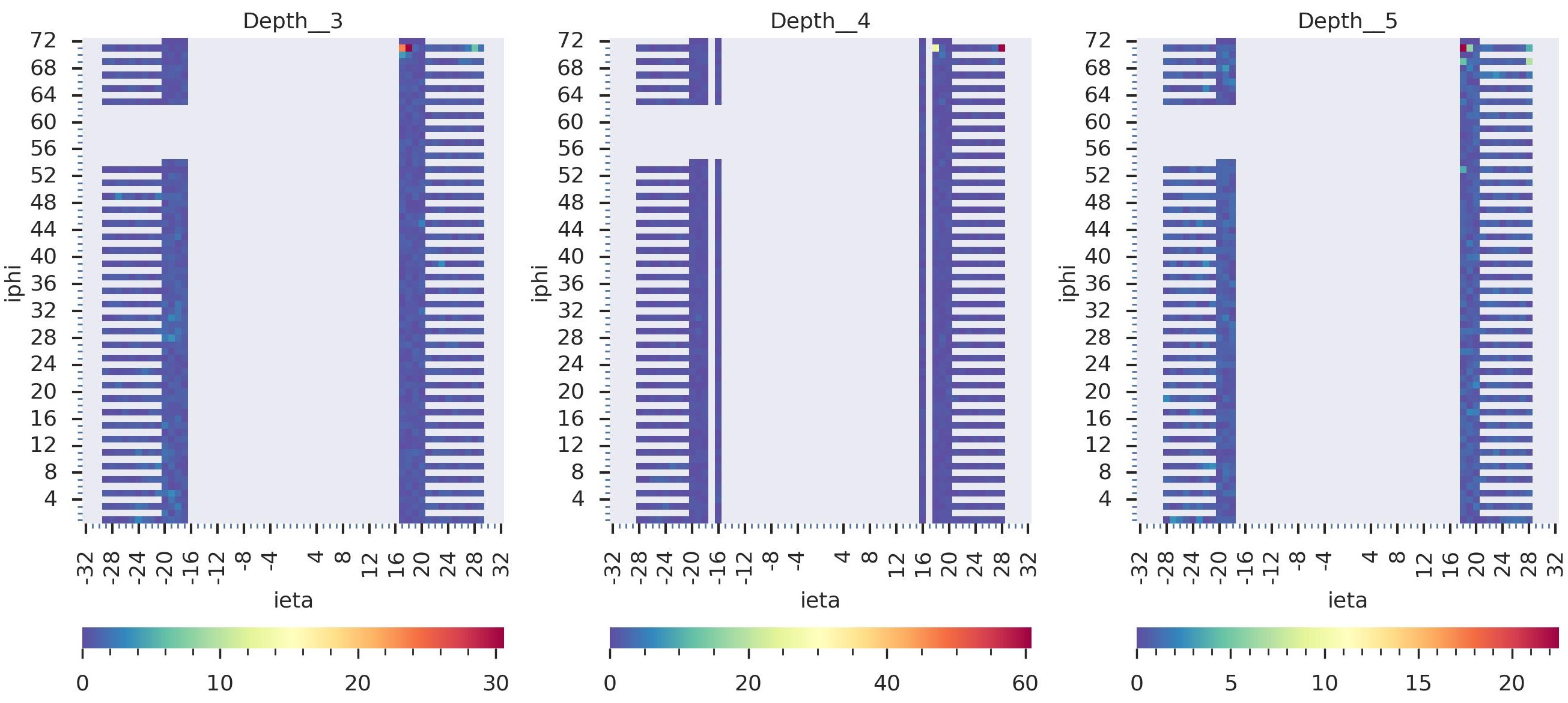}
\caption{}
\end{subfigure}
\caption{Spatial view on the detected real degraded channels at \textit{LS = 57} from \textit{RunId = 324841}. (\textbf{a}) The 2D digi-occupancy maps at the \textit{depth} axes of the faulty channels and (\textbf{b}) the corresponding anomaly score maps. The GraphSTAD localized the anomaly scores on the faulty degraded channels with a strength proportional to the anomaly severity---lower scores in the color bars than the dead channels.
}
\label{fig:digioccupancy_anomaly_hist2d_ls57}
\end{figure}

\subsection{Cost of Model Complexity}

\begin{figure}[H]
\includegraphics[width=0.6\textwidth, scale=1]{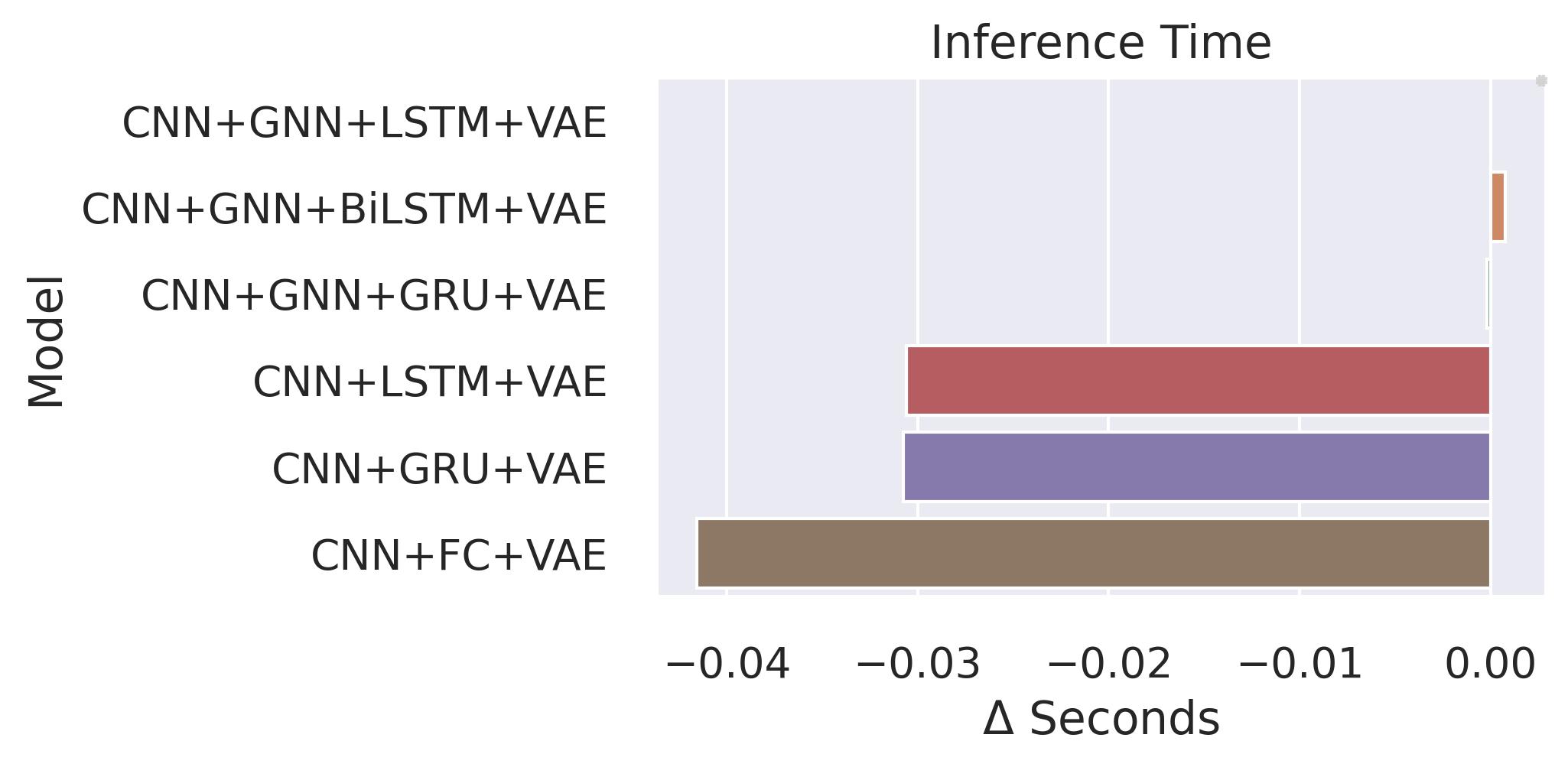}
\caption{Model inference computational cost relative to the proposed GraphSTAD model (CNN+GNN+LSTM+VAE). The GNN increased the inference delay, whereas the nontemporal model (CNN + FC + VAE) had a speed advantage due to its relatively lower number of model parameters and its inference on a single map instead of time windowing.}
\label{fig:inference_time_benchmark}
\end{figure}

We developed the models with PyTorch and trained them on four GPUs of NVIDIA Tesla V100 SXM3 32GB and an Intel(R) Xeon(R) Platinum 8168 CPU \@2.70 GHz. We utilized a time window $T=5$ and batch size $B=8$ for training, and the dimension of a batch was $[B \times T \times N_{i\eta} \times N_{i\phi} \times N_d \times N_f]$. The training time of the GraphSTAD model was approximately 45 s per epoch. 
The training iteration epoch 200 achieved good accuracy with a one-cycle learning rate schedule~\cite{smith2019super}. 
The nontemporal model---CNN + FC + VAE---was the fastest, and its superiority emanated from its nonrecurrent networks that only analyzed a single map instead of a sequential processing of five maps in a time window.
The median inference time of the GraphSTAD system on a single GPU was roughly 0.05 s with a standard deviation of 0.006 s. 
The integration of the GNN made the inference relatively slower compared to the benchmark models (see Figure \ref{fig:inference_time_benchmark}).
The processing cost was within an acceptable range for the CMS production requirement since the input digi-occupancy map was generated at each lumisection with a time interval of 23 s.

\section{Conclusions}
\label{sec:conclusion}

In this study, we presented a semi-supervised anomaly detection system for the data quality monitoring system of the {Hadron Calorimeter} using spatio-temporal digi-occupancy maps.
We extended the synergy of temporal deep learning developments for the CMS experiment. 
Our approach addressed modeling challenges, including digi-occupancy map renormalization, learning non-Euclidean spatial behavior, and degrading channel detection.
We proposed the GraphSTAD system that combined convolutional, graph, and temporal learning networks to capture spatio-temporal behavior and achieve a robust localization of anomalies at a channel granularity on high-dimensional spatial data. 
The AD performance evaluation demonstrated the efficacy of the proposed system for channel monitoring.
Our proposed AD system will facilitate monitoring and diagnostics of faults in the {front-end hardware and software systems of the calorimeter}. It will enhance the accuracy and automation of the existing DQM system, providing instant anomaly alerts on a broader range of channel faults in {realtime} and offline; the improved monitoring of the calorimeter will result in the collection of high-quality physics data.
The methods and approaches discussed in this study are domain-agnostic and can be adopted in other spatio-temporal fields, particularly when the data exhibit regular and irregular spatial characteristics.

\vspace{6pt}

\supplementary{The following supporting information can be {downloaded} at: \linksupplementary{s1}. Supplemental File: the CMS-HCAL {Collaboration}.}

\authorcontributions{Dataset curation, methodology, model development, and first draft preparation and reviewing, M.W.A.; methodology and supervision, C.W.O.; DQM data retrieval and preparation and model evaluation, L.W. and D.Y.; HCAL operations and read-out systems, P.P. and J.D.; ML methodology model development, and writing---first draft preparation and reviewing, M.W.A.;  
G.K., M.S., and R.V., collision data quality monitoring discussion;
L.L., E.U., M.A., J.F.M., and K.M., examination of the ML model design and performance evaluations; 
M.W.A., C.W.O., L.W., P.P., J.D., and R.V., editing and reviewing the manuscript; 
{The CMS-HCAL collaboration participated in maintaining the operations of CMS-HCAL and generating collision datasets, from which our datasets are derived.} 
All authors have read and agreed to the published version of the manuscript.}

\funding{{This research received no external funding.} 
}

\institutionalreview{{Not applicable.} 
}

\informedconsent{{Not applicable.} 
}

\dataavailability{ {Data are contained within the article.} 
}

\acknowledgments{{We} 
 sincerely appreciate the CMS collaboration, specifically the HCAL data performance group, the HCAL operation group, the CMS data quality monitoring groups, and the CMS machine learning core teams. Their technical expertise, diligent follow-up on our work, and thorough manuscript review have been invaluable.
We also thank the collaborators for building and maintaining the detector systems used in our study. We extend our appreciation to the CERN for the operations of the LHC accelerator. The teams at CERN have also received support from the Belgian Fonds de la Recherche Scientifique, and Fonds voor Wetenschappelijk Onderzoek; the Brazilian Funding Agencies (CNPq, CAPES, FAPERJ, FAPERGS, and FAPESP); SRNSF (Georgia); the Bundesministerium f\"ur Bildung und Forschung, the Deutsche Forschungsgemeinschaft (DFG), under Germany's Excellence Strategy--EXC 2121 ``Quantum Universe''---{390833306}, and under project number 400140256-GRK2497, and Helmholtz-Gemeinschaft Deutscher Forschungszentren, Germany; the National Research, Development and Innovation Office (NKFIH) (Hungary) under project numbers K~128713, K~143460, and TKP2021-NKTA-64; the Department of Atomic Energy and the Department of Science and Technology, India; the Ministry of Science, ICT and Future Planning, and National Research Foundation (NRF), Republic of Korea; the Lithuanian Academy of Sciences; the Scientific and Technical Research Council of Turkey, and Turkish Energy, Nuclear and Mineral Research Agency; the National Academy of Sciences of Ukraine; the US Department of Energy.
}

\conflictsofinterest{{The authors declare no conflict of interest} 
} 
\pagebreak
\abbreviations{Abbreviations}{
The following abbreviations are used in this manuscript:\\
\noindent 
\begin{tabular}{@{}ll}
    AE          &  Autoencoder \\ 
    AD          &  Anomaly detection \\ 
    CERN      & The European Organization for Nuclear Research \\
    CMS        &  Compact Muon Solenoid \\ 
    CNN         & Convolutional neural networks \\ 
    DL          & Deep learning \\
    DQM        &  Data quality monitoring \\ 
    FC        &  Fully connected neural networks \\ 
    GNN         & Graph neural networks \\ 
    GraphSTAD        &  Graph-based ST AD model \\ 
    HCAL        &  Hadron Calorimeter \\ 
    HE        &  HCAL Endcap detector \\ 
    HEP         &  High-energy physics \\
    LHC        &  Large Hadron Collider \\ 
    LS(s)      & Lumisection(s) \\
    MAE        &  Mean absolute error \\ 
    MSE        &  Mean square error \\
    QIE        & Charge integrating and encoding \\
    RBX        &  Readout box \\ 
    RNN         & Recurrent neural networks \\ 
    SiPM        & Silicon photomultipliers \\
    ST          & Spatio-temporal \\
    VAE          & Variational autoencoder \\

\end{tabular}
}

\begin{adjustwidth}{-\extralength}{0cm}
\reftitle{References}

\PublishersNote{}
\end{adjustwidth}
\end{document}